%% file: main.tex
\documentclass[10pt,twocolumn,letterpaper]{article}

\usepackage{cvpr}              % To produce the CAMERA-READY version
\usepackage{verbatim}
\usepackage{cite}
\usepackage{multirow}
\usepackage[numbers,sort&compress]{natbib}
\usepackage{url}
\usepackage{float}
\usepackage{bm}
\usepackage{graphicx}
\usepackage[accsupp]{axessibility} % Improves PDF readability for those with visual impairments.

\input{preamble}

\definecolor{cvprblue}{rgb}{0.21,0.49,0.74}
\usepackage[pagebackref,breaklinks,colorlinks,citecolor=cvprblue]{hyperref}

\def\paperID{7290} % *** Enter the Paper ID here
\def\confName{CVPR}
\def\confYear{2024}

\begin{document}

%%%%%%%%% TITLE - PLEASE UPDATE
\title{
    3D Building Reconstruction from Monocular Remote Sensing Images \\
    with Multi-level Supervisions
}

%%%%%%%%% AUTHORS - PLEASE UPDATE
\author{
    Weijia Li\textsuperscript{\rm 1}\footnotemark[1], 
    Haote Yang\textsuperscript{\rm 2}\footnotemark[1], 
    Zhenghao Hu\textsuperscript{\rm 1}, 
    Juepeng Zheng\textsuperscript{\rm 1}, 
    Gui-Song Xia\textsuperscript{\rm 3}, 
    Conghui He\textsuperscript{\rm 2,4}\footnotemark[2]
    \\
    \textsuperscript{\rm 1}Sun Yat-Sen University, 
    \textsuperscript{\rm 2}Shanghai AI Laboratory, 
    \textsuperscript{\rm 3}Wuhan University,
    \textsuperscript{\rm 4}SenseTime Research
    \\
    {\tt\small 
        \{liweij29, zhengjp8\}@mail.sysu.edu.cn, 
        \{yanghaote, heconghui\}@pjlab.org.cn,
    }
    \\
    {\tt\small
        huzhh9@mail2.sysu.edu.cn, 
        guisong.xia@whu.edu.cn
        %heconghui@sensetime.com
    } 
}

\maketitle

\renewcommand{\thefootnote}{\fnsymbol{footnote}}
\footnotetext[1]{These authors contributed equally to this work.}
\footnotetext[2]{Corresponding author.}

%input sections
\input{sec/0_abstract}    
\input{sec/1_intro}

\input{sec/2_related_work}

\input{sec/3_methods}
\input{sec/4_experiments}

\input{sec/5_conclusion}
\input{sec/6_suppl}

\clearpage
{
    \small
    \bibliographystyle{ieeenat_fullname}
    \bibliography{main}
}

\end{document}

%% file: preamble.tex
\usepackage[dvipsnames]{xcolor}

%% file: sec/0_abstract.tex
\begin{abstract}
%研究意义
3D building reconstruction from monocular remote sensing images is an important and challenging research problem that has received increasing attention in recent years, owing to its low cost of data acquisition and availability for large-scale applications.
%现有研究的不足
However, existing methods rely on expensive 3D-annotated samples for fully-supervised training, restricting their application to large-scale cross-city scenarios.
%总体介绍我们的方法
In this work, we propose MLS-BRN, a multi-level supervised building reconstruction network that can flexibly utilize training samples with different annotation levels to achieve better reconstruction results in an end-to-end manner. 
%具体有什么新模块，解决什么挑战
To alleviate the demand on full 3D supervision, we design two new modules, Pseudo Building Bbox Calculator and Roof-Offset guided Footprint Extractor, as well as new tasks and training strategies for different types of samples.
Experimental results on several public and new datasets demonstrate that our proposed MLS-BRN achieves competitive performance using much fewer 3D-annotated samples, and significantly improves the footprint extraction and 3D reconstruction performance compared with current state-of-the-art.
The code and datasets of this work will be released at \href{https://github.com/opendatalab/MLS-BRN.git}{https://github.com/opendatalab/MLS-BRN.git}.
\end{abstract}

% 3D building reconstruction from monocular remote sensing images is an important and challenging research problem that has received increasing attention in recent years, owing to its low cost of data acquisition and availability for large-scale applications. However, existing methods rely on expensive 3D-annotated samples for fully-supervised training, restricting their application to large-scale cross-city scenarios. In this work, we propose MLS-BRN, a multi-level supervised building reconstruction network that can flexibly utilize training samples with different annotation levels to achieve better reconstruction results in an end-to-end manner. To alleviate the demand on full 3D supervision, we design two new modules, Pseudo Building Bbox Calculator and Roof-Offset guided Footprint Extractor, as well as new tasks and training strategies for different types of samples. Experimental results on several public and new datasets demonstrate that our proposed MLS-BRN achieves competitive performance using much fewer 3D-annotated samples, and significantly improves the footprint extraction and 3D reconstruction performance compared with current state-of-the-art. The code and datasets of this work will be made publicly available.

%% file: sec/1_intro.tex
\section{Introduction}
\label{sec:intro}
%-------------------------------------------------------------------------

\begin{figure} [ht]
  \includegraphics[width=1.0\linewidth]{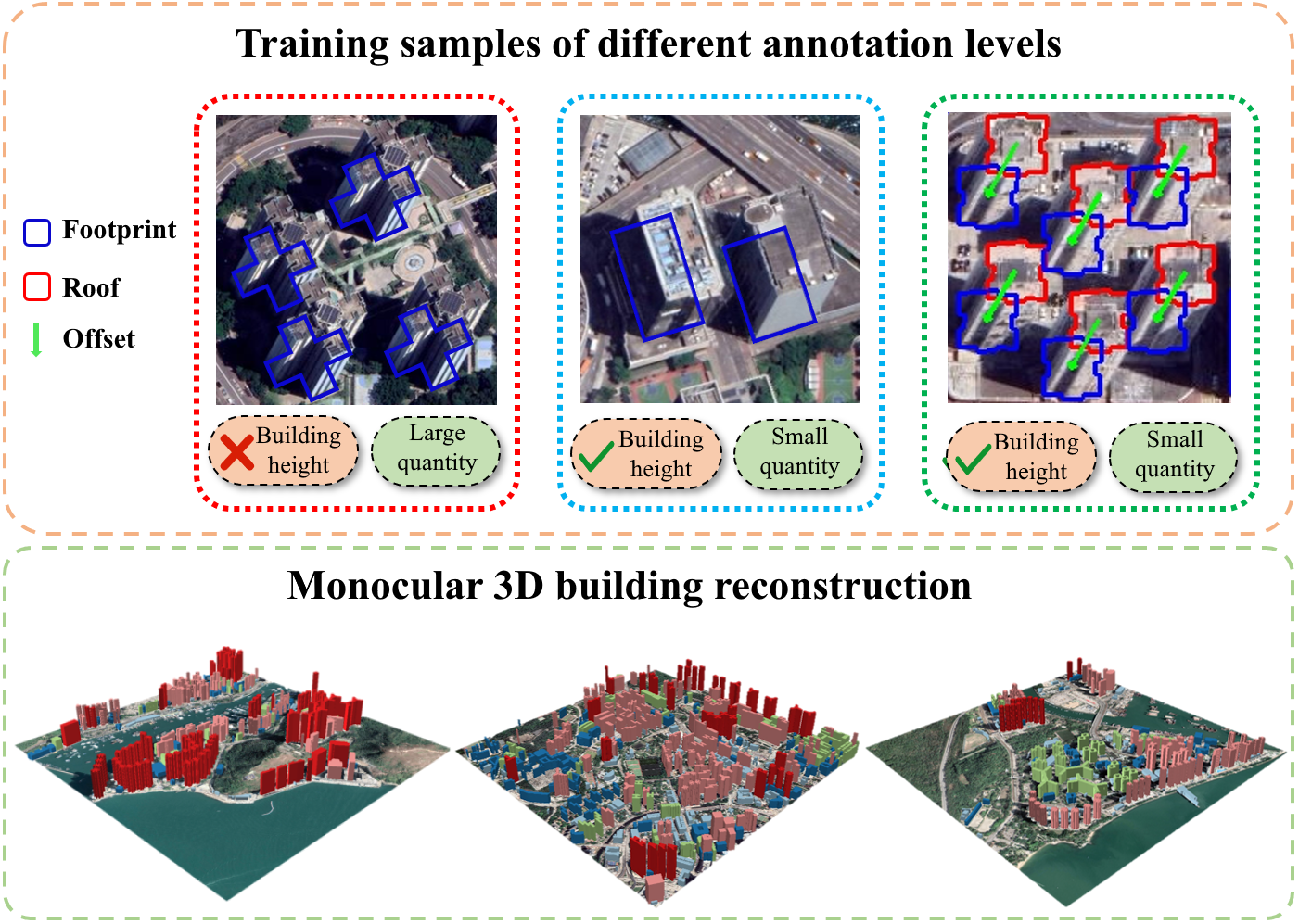}
  \caption{
  Our proposed method achieves 3D building reconstruction by training samples of different annotation levels.
  Large quantity of samples only include building footprint annotations, whereas a small quantity of samples contain extra roof-to-footprint offset and building height annotations.
  }
  \label{fig:intro}
\end{figure}

% 得益于较低的数据获取成本，单目重建已经受到越来越多的关注。

%para1: importance and challenge of monocular building reconstruction

3D building reconstruction is a fundamental task for large-scale city modeling and has received increasing attention in recent studies. Among these studies, monocular 3D building reconstruction has become a promising and economic solution for large-scale real-world applications, owing to its lower data acquisition cost and larger data coverage compared to multi-view stereo imagery and LiDAR data \cite{20063D,Duan2016Towards}.
Meanwhile, the limited information of monocular images as well as the diversity of building structures also result in great challenges for large-scale 3D building reconstruction.

% near-nadir related
Inspired by the progress of supervised monocular depth estimation methods, deep neural networks have been broadly applied to monocular 3D building reconstruction studies. 
Most studies utilize building footprints or other types of semantic labels as prior information to facilitate building height estimation from near-nadir images \cite{kunwar2019u,2020Boundary,2017Joint,zheng2019pop,mao2023elevation}.
%off-nadir Offset-related:
Off-nadir images, by contrast, constitute a larger proportion of the remote sensing images and provide additional useful information for building height estimation, which have demonstrated significant potential in several recent studies \cite{2020Learning,christie2021single,xiong2023benchmark,li20213d,wang2022learning}.
Some studies designed geocentric pose estimation task considering the parallax effect of building roof and footprint \cite{2020Learning,christie2021single}, aiming at estimating the height values instead of reconstruct a 3D model. Other studies leveraged the relation between different components of a building instance (e.g. roof, footprint, and facade) as well as the offset between roof and footprint, which has proven to be an effective solution for 3D building reconstruction and accurate extraction of building footprints \cite{li20213d,wang2022learning}.
% leveraging the relation between building components to
%and fail to explore the relation between different components of a building instance (e.g. roof, footprint, and facade), and the relation between the building heights and semantic types. 

In general, existing monocular building reconstruction methods are designed for fully-supervised learning, requiring a large number of fully-annotated 3D labels for network training.
However, due to the expensive annotation cost, the available datasets for 3D building reconstruction are still very insufficient, restricting existing 3D reconstruction methods to single city or single dataset scenarios. By contrast, owing to the low annotation cost and the increase of open map data, public building footprints have an extremely large coverage and quantity. Additionally, existing building datasets provide different levels of annotations, such as footprint only, footprint and pixel-wise height \cite{2020Learning}, footprint and offset vector \cite{li20213d,wang2022learning}, etc.
The large-scale 2D footprints and different levels of annotated datasets can provide new opportunities for enlarging 3D building reconstruction application scenarios and reducing the annotation cost if they are effectively utilized.

%%% our work
%% challenges (bbox, training),
In this work, we propose MLS-BRN, a Multi-Level Supervised Building Reconstruction Network based on monocular remote sensing images, which is a unified and flexible framework that is capable of utilizing the training samples with different annotation levels.
To alleviate the demand on 3D annotations and enhance the building reconstruction performance, we design new tasks regarding the meta information of off-nadir images and two new modules, i.e., Pseudo Building Bbox Calculator and Roof-Offset guided Footprint Extractor, as well as a new training strategy based on different types of samples.
%Specifically, new tasks regarding the meta information of off-nadir images and two novel modules (i.e., Pseudo Building Bbox Calculator and Roof-Offset guided footprint extractor) are designed to alleviate the demand on 3D annotations in existing fully-supervised reconstruction method.
Experimental results on several public and new datasets demonstrate that our method achieves competitive performance when only using a small proportion of 3D-annotated samples, and significantly improves the building segmentation and height estimation performance compared with current state-of-the-art. 
Our main contributions are summarized as follows:

\begin{itemize}

%%% version 1

%\item We design MLS-BRN, a multi-level supervised building reconstruction network, which is a unified and flexible framework that is capable of utilizing the training samples with different annotation levels.

%\item We design new tasks regarding the meta information of off-nadir images and new modules, i.e., Pseudo Building Bbox Calculator and Roof-Offset guided footprint extractor, effectively alleviating the demand on 3D annotations and enhancing the 3D reconstruction performance.

%%% version 2:

\item We design MLS-BRN, a multi-level supervised building reconstruction network, which consists of new tasks and modules to enhance the relation between different components of a building instance and alleviate the demand on 3D annotations.

\item We propose a multi-level training strategy that enables the training of MLS-BRN with different supervision levels to further improve the 3D reconstruction performance.

\item We extend the monocular building reconstruction datasets to more cities. Comprehensive experiments under different settings demonstrate the potential of MLS-BRN in large-scale cross-city scenarios.

% using: (1) entire 3D-annotated samples, (2) partial 2D-annotated samples + partial 3D-annotated labels, and (3) entire 3D-annotated samples + extra 2D-annotated samples. Results show that our method achieves competitive performance using fewer 3D labels, and significantly better performance using the same training set or extra 2D labels compared with current state-of-the-art methods.

%and enable the effective utilization of different supervision levels.

%with various types of tasks and two new modules, which 
%more angle-related and instance-wise height estimation, as well as two 
%fully explores the relation between the main components of a 3D building instance, achieving superior footprint segmentation and height estimation performance compared with current state-of-the-art methods.

%reduces the demand on large-scale training samples with expensive 3D annotations compared with the existing supervised building reconstruction methods.

\end{itemize}

%% file: sec/2_related_work.tex
\section{Related work}
\label{sec:related_work}
%-------------------------------------------------------------------------
\begin{figure*}
    \centering
    \includegraphics[width=0.95\linewidth]{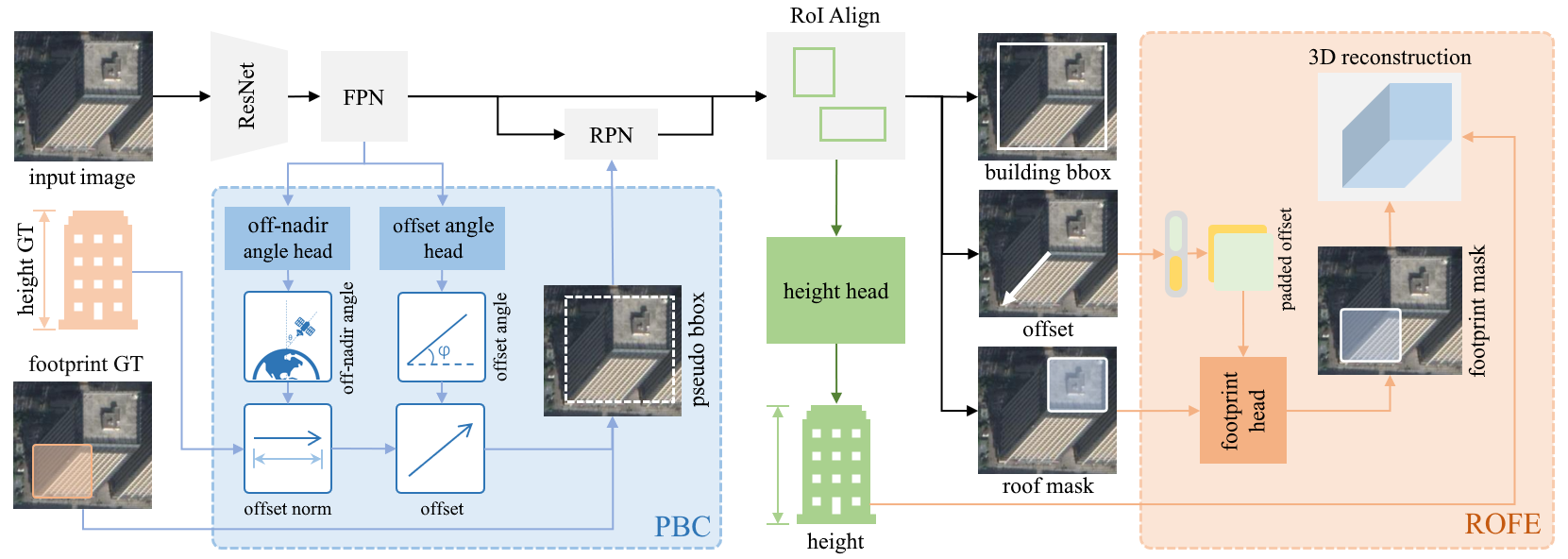}
    \caption{
    %The overall framework of our proposed MLS-BRN. 
    %Compared to LOFT-FOA, two new modules are added: the Roof-Offset guided Footprint Extractor (ROFE) and the Pseudo Building Bbox Calculator (PBC).
    %The backbone is the ResNet with Feature Pyramid Network (FPN). 
    %The RoI Align layers predict the building bbox, offset, roof mask, and building height.
    %The ROFE predicts footprint mask conditioned on the predicted roof mask and offset.
    %The predicted offset and footprint mask are used for the 3D reconstruction.
    %The PBC predicts off- nadir and offset angles and use them to calculate pseudo building bboxes for building bbox-unknown samples. 
    An overview of our proposed method. 
    Taking a monocular remote sensing image as input, our MLS-BRN generates a set of building bboxes, roof-to-footprint offsets, building heights, and pixel-wise roof masks.
    The predicted roof masks and their corresponding offsets are further integrated to predict pixel-wise footprint masks.
    The predicted footprint mask and building height are used to produce the final vectorized 3D model.
    Two novel modules are introduced: (1) the \textbf{ROFE} predicts footprint masks guided by the predicted roof masks and offsets; (2) the \textbf{PBC} predicts off-nadir and offset angles to calculate pseudo building bboxes for building bbox-unknown samples. 
    }
    \label{fig:overall-framework}
\end{figure*}

\subsection{Building footprint extraction}

% the pixel-wise segmentation methods, the contour-based methods, and the polygon-based methods.

Building footprint extraction is an important prerequisite for monocular 3D building reconstruction. Various instance and semantic segmentation networks have been broadly applied to building extraction tasks. 
Many studies utilize multi-task segmentation network to improve the building segmentation performance. For instance, Yuan \cite{yuan2017learning} proposed the signed distance representation for building footprint extraction, achieving better performance compared with the single-task fully-connected network.
%In \cite{bischke2019multi}, a multi-task learning method was proposed to improve the building boundary prediction performance, which introduced an extra task to predict the distance to the border of buildings using an encoder-decoder network architecture.  
Similarly, in \cite{2020Boundary}, a modified signed distance function was introduced and jointly learned with other tasks for predicting building footprint outlines and heights. 
To improve the geometry shapes of building extraction results, several methods directly predicted the vertices of a building polygon based on Recurrent Neural Network or Graph Neural Network \cite{li2019topological,zhao2021building,zorzi2022polyworld}, or combined the pixel-based multi-task segmentation network with a graph-based polygon refinement network using a rule-based module \cite{li2021joint}.
In addition, some recent studies converted building footprint extraction into roof segmentation and roof-to-footprint offset estimation tasks, which achieved promising performance for building footprint extraction, especially for high-rise buildings in off-nadir images \cite{li20213d,wang2022learning}.

In summary, most existing methods directly extract the building footprints and perform worse for high-rise buildings in off-nadir images. Offset-based methods can effectively alleviate this problem, but the expensive offset annotation efforts and the post-processing process are still inevitable.
On the contrary, our work proposes a multi-level supervised solution that is capable of leveraging different types of samples to reduce the demand for offset annotation, achieving promising footprint extraction results in an end-to-end manner.

%%%%%%%%%%%%%%
\subsection{Monocular 3D building reconstruction}

%There is an increasing number of studies for 3D building reconstruction from monocular remote sensing images, owing to the inexpensive data acquisition costs and broad data coverage compared with reconstruction from LiDAR data \cite{20063D} or multi-view imagery \cite{Duan2016Towards,2016Semantically,20183D,liu2020novel}. Traditional monocular 3D building reconstruction methods are mostly based on the shadow information, lines or line intersections of the building outlines, and the meta information of satellites such as the sun-earth relative position \cite{2012Three,2013Automated}. Complicated procedures with multiple steps are required for reconstructing the final 3D building model.

Inspired by the progress of monocular depth estimation, deep neural networks have been widely used for monocular building height estimation in recent studies \cite{gao2023joint,xiong2023benchmark,li20233dcentripetalnet}. Most of these studies are designed for height estimation from near-nadir images, in which the building roof and footprint are almost overlapped. Some methods used an encoder-decoder network to regress the height values \cite{mao2023elevation}, or used a generative adversarial network to simulate a height map \cite{2018IMG2DSM}. Moreover, the semantic labels have been utilized as effective priors in many existing methods considering the limited information provided from the near-nadir images for height estimation. Some studies designed a multi-task network for joint footprint extraction and height estimation \cite{2017Joint,zheng2019pop,gao2023joint}, while others exploit the semantic labels as prior information for height estimation \cite{kunwar2019u}.
In actual scenarios, off-nadir images constitute a large proportion of the remote sensing images, in which the parallax effect of roof and footprint results in more challenges for extracting footprints but provides additional information for height estimation as well.
Some recent studies \cite{2020Learning,christie2021single} design methods to learn the geocentric pose of buildings in off-nadir images for monocular height estimation \cite{ronneberger2015u}, while others leverage the offset between building roof and footprint and the relation between different components to reconstruct a 3D building model \cite{li20213d,wang2022learning}.
%and enhance building footprint extraction . 

In summary, the monocular building reconstruction methods in existing studies require expensive and fully-annotated 3D labels for supervised learning.
Our proposed method, by contrast, is a unified and flexible framework for 3D building reconstruction with different supervision levels, which effectively reduces the demand for the large-scale 3D annotations. 

\subsection{Monocular 3D reconstruction with fewer labels}

In monocular 3D reconstruction in the general computer vision domain, several methods have been proposed for reducing the 3D annotation demand via weakly-supervised or semi-supervised learning \cite{chen2019so,mitra2020multiview,han2021weakly,li2023robust,ji2019semi}. 
In Yang et al. \cite{yang2018learning}, a unified framework combining two types of supervisions was proposed, i.e., a small number of camera pose annotations and a large number of unlabeled images.
In Neverova et al. \cite{neverova2017hand}, an intermediate representation containing important topological and structural information of hand was introduced to enable the weakly-supervised training for hand pose estimation.
Concurrently, Gwak et al. \cite{gwak2017weakly} effectually leveraged a weak supervision type, i.e., foreground mask, as a substitute for costly 3D CAD annotations, which incorporates a raytrace pooling layer to enable perspective projection and backpropagation.

In contrast to the aforementioned studies, our proposed method leverages prior knowledge about the 3D structure of a  building instance and the monocular remote sensing image, including the relation between roof, footprint, height, offset angle, and off-nadir angle, enabling multi-level supervised 3D reconstruction with fewer annotation efforts.

%% file: sec/3_methods.tex
\section{Methods}
\label{sec:methods}
% ============================================ Problem Statement ============================================
%%lwj: 如果整体框架图3是3.2小节才提到的话，3.1小节应该也放一个示例图。现在3.1还是文字太多，文字公式也看不出这几个task之间的关系。
\subsection{Problem statement}
% 描述任务，主要挑战，前人典型做法
Given an off-nadir remote sensing image $I$ that includes buildings $B=\{b_1, b_2, ..., b_N\}$, the objective of monocular 3D building reconstruction is to identify all the footprints $F=\{f_1, f_2, ..., f_N\}$ and roofs $R=\{r_1, r_2, ..., r_N\}$ corresponding to $B$. 
The difficulty is that the footprints of buildings may be partially visible from an off-nadir viewing angle. 
Thus, previous studies, including \cite{li20213d} and \cite{wang2022learning}, typically solve this issue by training a deep neural network with samples annotated with both $F$ and roof-to-footprint offsets $\vec{V}=\{v_1, v_2, ..., v_N\}$.  

% 强调标注的稀缺性，引出我们的解决方案
However, the cost of annotating remote sensing images is still high, particularly for offset annotations.
Therefore, we suggest addressing this issue by training a deep model that effectively uses samples containing both $F$ and $\vec{V}$ annotations, alongside samples only annotated with $F$.

% 我们的方案新增加的任务
To facilitate training with offset-unknown samples, two tasks are included; one for predicting the off-nadir angle $\theta_{I}$ and the other for the offset angle $\varphi_{I}$. 
Additionally, an instance-wise footprint segmentation task is included to predict the footprint conditioned on the predicted roof and offset.
Finally, a task for predicting real-world height is introduced to enhance the comprehension of the correlation between footprint and roof placement.
In summary, four additional tasks are added to the original three tasks in LOFT-FOA \cite{wang2022learning}: (1) off-nadir angle prediction task; (2) offset angle prediction task; (3) footprint segmentation task; (4) real-world height prediction task.

% =============================================== Structure ===============================================
\subsection{Network structure}
% 介绍MLS-BRN在loft-foa的基础上增加的三个模块
\cref{fig:overall-framework} illustrates the proposed architecture of our MLS-BRN.
To facilitate multi-level supervised learning, two novel modules are introduced, namely the Pseudo Building Bbox Calculator (PBC) and the Roof-Offset guided Footprint Extractor (ROFE). 
% 介绍增加的三个新增加模块的功能
The PBC module provides pseudo building boxes to determine the positivity/negativity of the region proposals from the RPN module when offset-unknown (\ie building bbox-unknown) samples are processed in the MLS-BRN. 
The ROFE module has two significant functions. 
Firstly, it provides a more straightforward method to supervise the building footprint segmentation task.
Secondly, it offers an indirect method of supervising offset prediction and roof segmentation for offset-unknown samples as they pass through the MLS-BRN.
Additionally, a building height prediction task has been included in order to predict the real-world building height.

\subsubsection{Pseudo Building Bbox Calculator (PBC)}
% 介绍PBC模块的功能
%%lwj-done: 上面这句话没啥用。motivation写成一段就行，现在写的太朴实了，没把这个方法的巧妙性写出来，要写我们怎么设计的，怎么根据这几个我们新加的预测任务获得psudeo building bbox。
Samples without the ground truth for building bounding box $b\text{-}bbox_{gt}$ cannot be utilized by previous models, like LOFT-FOA \cite{wang2022learning}.
To address this issue, we propose a module that predicts pseudo building bounding boxes to substitute $b$-$bbox_{gt}$.
For a provided off-nadir remote sensing image $I$ and one building $b$ contained by $I$, we can describe the connection between the image-wise off-nadir angle $\theta_{I}$, the offset angle $\varphi_{I}$, the factor for scaling real-world height to pixel scale $s_I$, and the building's height $h_b$ and offset $\vec{v}_b$ using the following equation:
\begin{equation}
\begin{aligned}
\vec{v}_b   &= ||\vec{v}_b||_2 \times \vec{e}   \\ \label{eq:v}
            &= ||\vec{v}_b||_2 \times [e_x, e_y]   \\
            &= h_b \times s_I \times \tan{\theta_I} \times [\cos{\varphi_I}, \sin{\varphi_I}]
\end{aligned}
\end{equation}
where $||\vec{v}_b||_2$ is the $L2$ norm of the offset, $\vec{e}$ is the unit normal vector of $\vec{v}_b$.
%, $v_x$ and $v_y$ are the components of $\vec{v}$ on the x-axis and y-axis.
%So if we can get a $\theta_{I}$ and a $\varphi_{I}$, we can calculate a building bbox.
%%lwj-done: 这句子写的。。太重复了。而且到底哪个是image-wise
The PBC module uses an off-nadir angle head to predict an image-wise off-nadir angle $\theta_{pred}$ and an offset angle head to predict an image-wise offset angle $\varphi_{pred}$. 
Then, following \cref{eq:v}, they are combined with the instance-wise building height ground truth $h_{gt}$, and scale factor $s_{gt}$ to compute the pseudo offset $\vec{v}_{pred}$.
Finally, $f_{gt}$ is translated to get the pseudo building bbox $b\text{-}bbox_{pred}$ guided by $\vec{v}_{pred}$.
$b$-$bbox_{pred}$ will play the role of $b$-$bbox_{gt}$ during the training of the building bbox-unknown samples.
% 从弱监督角度进行简单解释
From the perspective of weak supervision, the PBC module extracts the image-wise angle information, \ie the offset angle and the off-nadir angle, and uses it to supervise the instance-wise task.
% 解释无高度弱监督的样本的假框如何计算
Note that for building height-unknown samples, the pseudo bounding boxes are calculated by directly enlarge the footprint boxes.

\subsubsection{Roof-Offset guided Footprint Extractor (ROFE)}
% 介绍ROFE模块的功能
%The ROFE module directly predicts the footprint mask.
Previous works calculate the footprint mask in the inference stage by translating the inferred roof guided by the inferred offset.
The ROFE module, however, predicts the footprint mask directly.
It trains a convolutional network to learn the translation process, using the inferred roof mask and offset as inputs.
For offset-aware (\ie roof-aware) samples, this end-to-end training process adds more supervision on the offset head and the roof head. 
And for offset-unknown samples, which cannot contribute to the training of the offset head and the roof head due to lack of ground truth, ROFE provides an indirect way to supervise these two heads.

% ================================================ Training ================================================
\subsection{Network training}
% 介绍本小节主要内容
In this section, we first introduce the loss functions in our MLS-BRN.
Then we introduce our three levels of training samples graded by their level of supervision and their training strategies.
The total hybrid loss is presented at the end of this section. 

%%lwj-done: 小标题起得都不是很好。应该3.3.1是任务相关的loss定义，3.3.2是不同层级监督训练。现在的小标题和里面的内容不搭，也没和主要贡献点对应上。
\subsubsection{Loss definition}
% 介绍loft-foa的损失
The LOFT-FOA \cite{wang2022learning} is trained by minimising \cref{eq:loft-foa-loss}, where $\mathcal{L}_{rp}$, $\mathcal{L}_{rc}$, $\mathcal{L}_{mh}$ are the same as those in Mask R-CNN \cite{he2017mask}, \ie, the losses for the RPN, R-CNN, and mask head, respectively; $\mathcal{L}_{o}$ is the loss for the offset head, which is a standard smooth L1 Loss.
\begin{equation}
\mathcal{L}_{LF} = \mathcal{L}_{rp} + \beta_1\mathcal{L}_{rc} + \beta_2\mathcal{L}_{mh} + \beta_3\mathcal{L}_{o}  \label{eq:loft-foa-loss}
\end{equation}

% 承上启下
The MLS-BRN model keeps the four losses the same as LOFT-FOA \cite{wang2022learning} and introduces new losses to train the newly added modules.
% ROFE模块的损失
The footprint mask loss of the ROFE module is the same as $\mathcal{L}_{mh}$, which is a standard cross entropy loss (\cref{eq:f-mask-loss}).
\begin{equation}
    \mathcal{L}_{f} = \frac{1}{N}\sum^{N}_{i=1}\sum^{C}_{c=1}y_{i,c} \times \log{(p(y_{i, c}))} \\ \label{eq:f-mask-loss}
\end{equation}

% PBC模块偏移角度头的损失
The loss of the offset angle head of the PBC module is calculated according to \cref{eq:ova-loss}, in which $\mathcal{L}_{ova}$ denotes the offset angle loss; $\vec{v}_{pred}$ denotes the predicted unit normal vector of the offset.
\begin{equation}
\begin{aligned}
\mathcal{L}_{ova}   &= \mathcal{L}_{ang} + \lambda_1\mathcal{L}_{reg} \label{eq:ova-loss}   \\
                    &= ||\vec{v}_{pred} - \vec{v}_{gt}||_1 + \lambda_1|| ||\vec{v}_{pred}||_2 - 1||_1   \\
\end{aligned}
\end{equation}

% PBC模块天顶角度头的损失
The nadir angle head of the PBC module is trained following \cref{eq:ona-loss}, where $\mathcal{L}_{ona}$ is the off-nadir angle loss; $\theta_{pred}$ is the predicted tangent of the off-nadir angle.
\begin{equation}
\mathcal{L}_{ona} = ||\tan{\theta}_{pred} - \tan{\theta}_{gt}||_1     \label{eq:ona-loss}   \\
\end{equation}

% 高度头的损失
The height head loss of our MLS-BRN is calculated by \cref{eq:h-loss}, in which $\mathcal{L}_{h}$ denotes the height loss; $h_{pred}$ denotes the predicted building height.
\begin{equation}
\mathcal{L}_{h} = ||h_{pred} - h_{gt}||_1     \label{eq:h-loss}   \\
\end{equation}

%%lwj-done: 超参数设置（loss权重）放到3.4里，不能empirically，要写一点原因，比如哪些任务是比较重要的。
\subsubsection{Multi-level training strategy}
% 三个层次的数据集
In our proposed unified framework, all the training samples can be graded into three levels according to their level of supervision (\cref{fig:intro}):
\begin{itemize}
    \item Level 1 samples: samples with only instance-wise footprint annotation, which are denoted by $\mathcal{X}^{N} = \{ x^{N}_1, x^{N}_2, ..., x^{N}_{n_3} \}$. $N$ means no additional supervision.
    \item Level 2 samples: samples with instance-wise footprint and building height annotation, which are denoted by $\mathcal{X}^{H} = \{ x^{H}_1, x^{H}_2, ..., x^{H}_{n_2} \}$.
    \item Level 3 samples: samples with instance-wise footprint, offset, and building height annotation, which are denoted by $\mathcal{X}^{OH} = \{ x^{OH}_1, x^{OH}_2, ..., x^{OH}_{n_1} \}$. 
\end{itemize}

% 承上启下
Different levels of samples are supervised by different training strategies.
% 只有底座的样本的损失
%The loss function for $\mathcal{X}^{N}$ is defined in \cref{eq:x-n-loss}. The only activated loss for $\mathcal{X}^{N}$ is $\mathcal{L}_{f}$.
As defined in \cref{eq:x-n-loss}, the loss function for $\mathcal{X}^{N}$ is only based on $\mathcal{L}_{f}$.
\begin{equation}
    \mathcal{L}_{\mathcal{X}^{N}} = \mathcal{L}_{f}   \\ \label{eq:x-n-loss}
\end{equation}

% 有底座和高度的样本的损失
The loss function for $\mathcal{X}^{H}$ is defined in \cref{eq:x-h-loss}.
In $\mathcal{L}_{\mathcal{X}^{H}}$, the $\mathcal{L}_{rp}$ is activated since the PBC module can predict a high-quality pseudo building bbox, which is good enough to supervise the RPN module.
%However, the pseudo building bbox is not accurate enough for training $\mathcal{L}_{R-CNN}$, $\mathcal{L}_{mh}$, and $\mathcal{L}_{Offset}$.
\begin{equation}
\begin{aligned}
    \mathcal{L}_{\mathcal{X}^{H}}   &= \mathcal{L}_{\mathcal{X}^{N}} + \alpha_1\mathcal{L}_{rp} + \alpha_2\mathcal{L}_{h}  \\
                                    &= \mathcal{L}_{f} + \alpha_1\mathcal{L}_{rp} + \alpha_2\mathcal{L}_{h} \label{eq:x-h-loss}
\end{aligned}
\end{equation}

% 有偏移的样本的损失
The loss function for $\mathcal{X}^{OH}$ is defined in \cref{eq:x-oh-loss}.
Compared with the original $\mathcal{L}_{LF}$, $\mathcal{L}_{\mathcal{X}^{OH}}$ adds four more losses: $\mathcal{L}_{f}$, $\mathcal{L}_{h}$, $\mathcal{L}_{ona}$, $\mathcal{L}_{ova}$. 
The $\mathcal{L}_{ona}$ and $\mathcal{L}_{ova}$ are used for training the two angle heads of the PBC module.
\begin{equation}
\begin{aligned}
    \mathcal{L}_{\mathcal{X}^{OH}}  =& \mathcal{L}_{\mathcal{X}^{H}} + \alpha_3\mathcal{L}_{rc} + \alpha_4\mathcal{L}_{mh}   \\ \label{eq:x-oh-loss} 
                                    +& \alpha_5\mathcal{L}_{o} + \alpha_6\mathcal{L}_{ona} + \alpha_7\mathcal{L}_{ova}    \\  
                                    =& \mathcal{L}_{LF} + \mathcal{L}_{f} + \alpha_2\mathcal{L}_{h} + \alpha_6\mathcal{L}_{ona} + \alpha_7\mathcal{L}_{ova}
\end{aligned}
\end{equation}

% 总的损失
The final hybrid loss is defined as the total loss of the three levels of training samples according to \cref{eq:overall-loss}. 
\begin{equation}
    \mathcal{L} = \mathcal{L}_{\mathcal{X}^{N}} + \mathcal{L}_{\mathcal{X}^{H}} +  \mathcal{L}_{\mathcal{X}^{OH}} \\ \label{eq:overall-loss}
\end{equation}

% ======================================== Implementation details ==========================================
\subsection{Implementation details}
As mentioned in \cref{fig:overall-framework}, we use ResNet-50 \cite{2016deep} with FPN \cite{lin2017feature} pre-trained on the ImageNet as the backbone. 
All the models are trained with a batch size of 4 using NVIDIA 3090 GPUs. 
To align with LOFT-FOA \cite{wang2022learning}, we train 24 epochs for all the models, with the learning rate starting from 0.01 and decaying by a factor of 0.1 at the $16^{th}$ and $22^{nd}$ epochs.
The SGD algorithm with a weight decay of 0.0001 and a momentum of 0.9 is used for all experiments.
LOFT-FOA \cite{wang2022learning} is used as the basic architecture of the MLS-BRN model, and all the hyperparameters that occur in both LOFT-FOA \cite{wang2022learning} and MLS-BRN are the same, except for the learning rate mentioned above.
All models are built in PyTorch.

In \cref{eq:ova-loss}, we set $\lambda_1 = 0.1$ to balance the two loss items.
In \cref{eq:x-h-loss}, we set $\alpha_1 = 1$ to keep the loss weight of ROFE the same as the roof mask head, and set $\alpha_2 = 32$ since the absolute building height loss value is relatively small.
In \cref{eq:x-oh-loss}, we set $\alpha_3 = \alpha_4 = 1, \alpha_5 = 16$ to keep them the same as LOFT-FOA \cite{wang2022learning}, and set $\alpha_6 = 1, \alpha_7 = 8$ to balance the effects of the magnitude of these two losses.

%% file: sec/4_experiments.tex
\section{Experiments}
\label{sec:experiments}

% ============================================ Datasets ===========================================
\subsection{Datasets}
% 一小段介绍数据集的内容
In our experiments, we employ multi-supervised datasets for training our methods: (1) \textbf{BONAI} \cite{wang2022learning} provides building footprint segmentation, offset, and height annotations, which contains 3,000 and 300 images for train-val and test respectively; (2) \textbf{OmniCity-view3} \cite{li2023omnicity} originally provides satellite images with annotations for footprint segmentation and building height. We add additional offset annotations for 17,092 and 4,929 images from train-val and test sets respectively; (3) Additionally, we release a new dataset named \textbf{HK}, which includes 500 and 119 satellite images specifically captured from Hong Kong for train-val and test sets, along with annotations for footprint segmentation, offset and height.

% 说明数据集的符号系统
As detailed in \cref{sec:methods}, all our training samples are graded into three levels: samples from $\mathcal{X}^{N}$, $\mathcal{X}^{H}$, and $\mathcal{X}^{OH}$.
To create different levels of training samples, we extract samples from the datasets mentioned above, reorganizing their annotations as necessary.
We randomly choose 30\% of the samples from the BONAI dataset \cite{wang2022learning} as a smaller $\mathcal{X}^{OH}$ dataset, which we call $BN_{30}$.
We randomly drop the offset annotations of 70\% of the samples in the BONAI dataset \cite{wang2022learning}, regard the entire BONAI \cite{wang2022learning} dataset as a $\mathcal{X}^{OH}$+$\mathcal{X}^{H}$ dataset, and name it $BN_{30/70}$.
Similarly, the original BONAI dataset \cite{wang2022learning} is regarded as a large $\mathcal{X}^{OH}$ and is named $BN_{100}$.
We use $OC$ to designate the OmniCity-view3 dataset \cite{li2023omnicity}. Naturally, the abbreviations $OC_{30}$, $OC_{30/70}$, and $OC_{100}$ have the similar meaning with $BN_{30}$, $BN_{30/70}$, and $BN_{100}$ respectively.
Moreover, we use $BH$ to refer to the combination of BONAI \cite{wang2022learning} and HK.
It is important to note that in $BH_{30/70}$, 30\% of BONAI's \cite{wang2022learning} samples are $\mathcal{X}^{OH}$ type while the remaining 70\% are $\mathcal{X}^{H}$ type.
Additionally, 30\% of HK's samples belong to $\mathcal{X}^{OH}$ type and the remaining 70\% belong to $\mathcal{X}^{N}$ type.

% ========================================== Performance ==========================================
\subsection{Performance comparison}
% 介绍章节内容
In this section, we evaluate our method's performance in footprint segmentation, offset prediction, and height prediction against several competitive methods for the single-level supervised learning scenario. 
In a Multi-level supervised learning scenario, we mainly compare our method with LOFT-FOA \cite{wang2022learning}.
Additionally, we present our method's offset and off-nadir angles prediction performance.
More results will be provided in the supplementary materials.

\begin{figure}[H]\centering
    \includegraphics[width=1.0\linewidth]{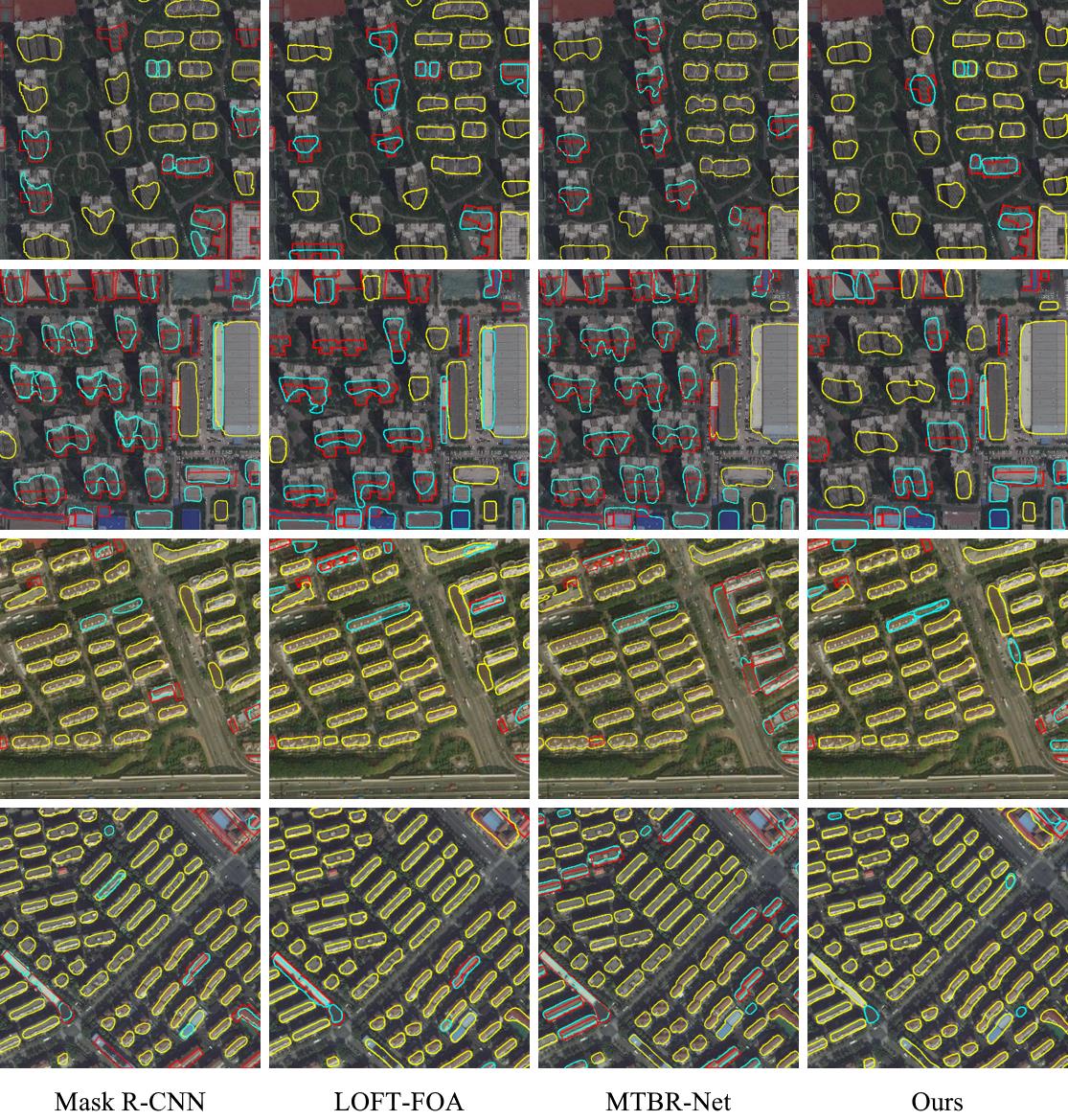}
    \caption{
    The results of the baselines and our method trained on $BN_{100}$ and tested on the BONAI test set in terms of the footprint segmentation performance. 
    The yellow, cyan, and red polygons denote the TP, FP, and FN.
    }
    \label{fig:footprint-seg}
\end{figure}

% 单层级监督学习
\textbf{Single-level supervised learning.} The performance of footprint segmentation and offset prediction for different methods trained on $BN_{100}$ and $OC_{100}$ are listed in \cref{tab:seg-offset-perf-bn} and \cref{tab:seg-offset-perf-oc}, respectively.
Additionally, \cref{fig:footprint-seg} provides a qualitative comparison of footprint segmentation results on the BONAI \cite{wang2022learning} test set.
Note that all the experimental results in this section are obtained using $\mathcal{X}^{OH}$ samples, and the results obtained using $\mathcal{X}^{H}$ and $\mathcal{X}^{N}$ samples will be analysed in the following paragraph. 
% 实验结果
For the footprint segmentation task, experimental results tested on $BN_{100}$ demonstrate that our method improves the F1-score by 5.42\% - 8.30\% compared with the instance segmentation methods that directly extract the building footprints.
Furthermore, our method enhances the F1-score by 2.05\% - 2.76\% relative to MTBR-Net \cite{li20213d} and LOFT-FOA \cite{wang2022learning}, which are specifically designed for extracting off-nadir building footprints based on predicted roof and offset, tested on $BN_{100}$.
Regarding the offset prediction task, our experimental findings indicate that our approach betters the EPE by 0.18 - 0.93 in comparison to MTBR-Net \cite{li20213d} and LOFT-FOA \cite{wang2022learning} tested on $BN_{100}$.
The results show that the direct supervision of the footprint segmentation, the constraint on the building height, and the encouragement of the angular feature extraction can help to achieve better performance in the footprint segmentation and offset prediction tasks in the single-level supervised learning scenario.

%单层级监督分割和偏移指标BONAI
\begin{table}[H]
    \centering
    \scalebox{0.8}{
    \begin{tabular}{lcccccccccc}
        \toprule
        method              & F1            & Precision     & Recall        & EPE 
        \\
        \midrule
        %Cascade MR-CNN      & 56.48         & 56.63         & 56.32         & -          
        PANet \cite{2020Approximating}        & 58.06         & 59.26         & 56.91         & -             
        \\
        HRNetv2 \cite{sun2019high_arxiv}      & 60.81         & 61.20         & 60.42         & -             
        \\
        M R-CNN \cite{he2017mask}             & 58.12         & 59.26         & 57.03         & -             
        \\
        CM R-CNN \cite{HTC_2019_CVPR}         & 60.94         & \textbf{67.09}   & 55.83       & -             
        \\
        MTBR-Net \cite{li20213d}              & 63.60         & 64.34         & 62.87         & 5.69          
        \\
        LOFT-FOA \cite{wang2022learning}     & 64.31         & 63.37         & 65.29           & 4.94          
        \\
        Ours                      & \textbf{66.36}        & 65.90    & \textbf{66.83}       & \textbf{4.76} 
        \\
        \bottomrule
    \end{tabular}
    }
    \caption{
    %The results of the baselines and our method trained on $BN_{100}$ and tested on the BONAI test set in terms of the footprint segmentation and offset prediction performance. 
    Building footprint segmentation results of different methods in terms of F1-score, precision, recall (\%) and offset prediction results in terms of EPE trained on $BN_{100}$.
    }
    \label{tab:seg-offset-perf-bn}
\end{table}

%单层级监督分割和偏移指标OmniCity-view3
\begin{table}[H]
    \centering
    \scalebox{0.95}{
    \begin{tabular}{lcccccccccc}
        \toprule
        method      & F1            & Precision     & Recall        & EPE           
        \\
        \midrule
        M R-CNN \cite{he2017mask}  & 69.75         & \textbf{69.74}         & 69.76         & -             
        \\
        LOFT-FOA \cite{wang2022learning}    & 70.46         & 68.77         & 72.23         & 6.08          
        \\
        Ours        & \textbf{72.25}& 69.57 & \textbf{75.14} & \textbf{5.38} 
        \\
        \bottomrule
    \end{tabular}
    }
    \caption{
    %The quantitative comparison of the baselines and our method trained on $OC_{100}$ and tested on the OmniCity-view3 test set in terms of the footprint segmentation and offset prediction performance. 
    Building footprint segmentation results of different methods in terms of F1-score, precision, recall (\%) and offset prediction results in terms of EPE trained on $OC_{100}$.
    }
    \label{tab:seg-offset-perf-oc}
\end{table}

% 多层级监督分割和偏移指标
\begin{table}[H]
    \centering
    \scalebox{0.83}{
    \begin{tabular}{llcccc}
        \toprule
        method      & dataset       &sample     & F1-score      & EPE           
        \\
        \midrule    
        LOFT-FOA \cite{wang2022learning}    & $BN_{30}$     &$\mathcal{X}^{OH}$ & 61.35          & 5.70          
        \\
        Ours        & $BN_{30/70}$  &$\mathcal{X}^{OH}$+$\mathcal{X}^{H}$ & 65.49          & 5.39          
        \\
        LOFT-FOA \cite{wang2022learning}    & $BN_{100}$    &$\mathcal{X}^{OH}$ & 64.31          & 4.94          
        \\
        Ours        & $BN_{100}$    &$\mathcal{X}^{OH}$ & 66.36          & 4.76          
        \\
        \midrule
        LOFT-FOA \cite{wang2022learning}    & $OC_{30}$     &$\mathcal{X}^{OH}$ & 67.09          & 6.08          
        \\
        Ours        & $OC_{30/70}$  &$\mathcal{X}^{OH}$+$\mathcal{X}^{H}$ & 70.53          & 5.92          
        \\
        LOFT-FOA \cite{wang2022learning}    & $OC_{100}$    &$\mathcal{X}^{OH}$ & 70.46          & 5.38          
        \\
        Ours        & $OC_{100}$    &$\mathcal{X}^{OH}$ & 72.25          & 5.38          
        \\
        \midrule
        LOFT-FOA \cite{wang2022learning}    & $BH_{30}$     &$\mathcal{X}^{OH}$ & 54.96          & 5.78          
        \\
        Ours        & $BH_{30/70}$  &$\mathcal{X}^{OH}$+$\mathcal{X}^{H}$+$\mathcal{X}^{N}$ & 58.57          & 5.60          
        \\
        LOFT-FOA \cite{wang2022learning}    & $BH_{100}$    &$\mathcal{X}^{OH}$ & 60.85          & 4.74          
        \\
        Ours        & $BH_{100}$    &$\mathcal{X}^{OH}$ & 60.92          & 4.69          
        \\
        \bottomrule
    \end{tabular}
    }
    \caption{
    %The quantitative comparison of the baselines and our method trained on different datasets in terms of the footprint segmentation and the offset prediction performance.
    Building footprint segmentation results of different methods in terms of F1-score (\%) and offset prediction results in terms of EPE trained on different datasets.
    }
    \label{tab:seg-offset-perf-wsl}
\end{table}

% 多层级监督学习
\textbf{Multi-level supervised learning.} \cref{tab:seg-offset-perf-wsl} displays the footprint segmentation and offset prediction performance of LOFT-FOA \cite{wang2022learning} and our method when trained and tested on multi-level supervision datasets.
Our approach's experiment outcomes, trained on $BN_{30/70}$, $OC_{30/70}$ and $BH_{30/70}$, demonstrate a 4.14\%, 3.44\% and 3.61\% improvement in F1-score compared to LOFT-FOA \cite{wang2022learning} trained on $BN_{30}$, $OC_{30}$ and $BH_{30}$.
Additionally, our method's experimental results, trained on samples from $BN_{30/70}$, $OC_{30/70}$ and $BH_{30/70}$ exhibit similar performance to LOFT-FOA \cite{wang2022learning}, which is trained on samples from $BN_{100}$, $OC_{100}$ and $BH_{100}$.
These findings demonstrate the effectiveness of MLS-BRN in combining samples from $\mathcal{X}^{OH}$, $\mathcal{X}^{H}$ and $\mathcal{X}^{N}$ levels to address the building reconstruction task.
% \begin{figure*}
%     \centering
%     \includegraphics[width=1.0\linewidth]{figures/3D-vis-OmniCity.png}
%     \caption{
%     The 3D reconstruction results on the OmniCity-view3 test set.
%     }
%     \label{fig:3d-recon}
% \end{figure*}

\begin{figure}[H]
    \centering
    \includegraphics[width=0.75\linewidth]{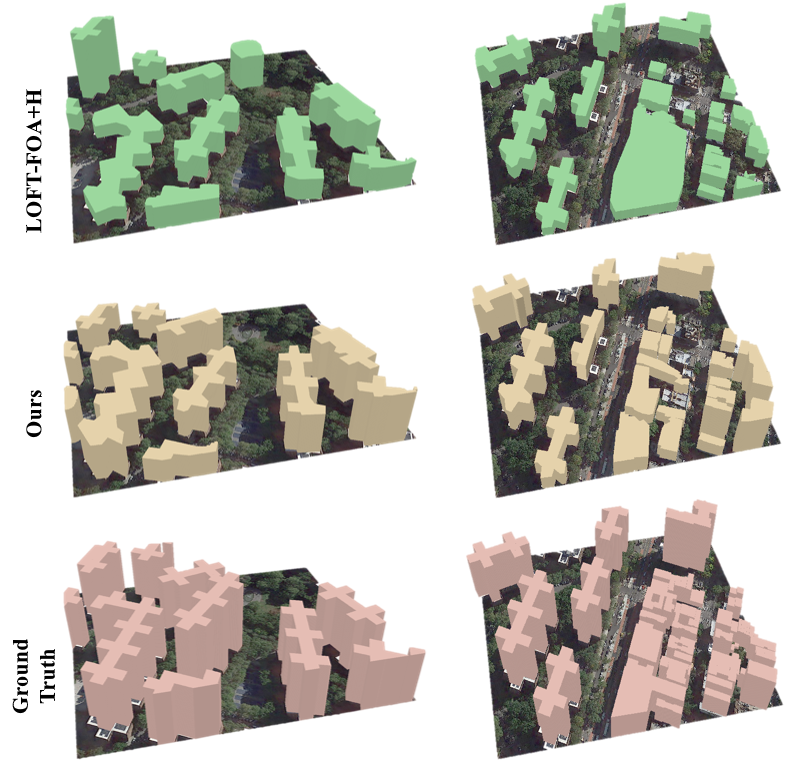}
    \caption{
    The visualization results of building height prediction from our method and LOFT-FOA+H on the OmniCity-view3 test set.
    }
    \label{fig:building-height}
\end{figure}

% 高度和角度指标
\textbf{Building height and angles prediction.} \cref{tab:height-perf} displays the results of building height prediction performance.
The experimental findings indicate that our method enhances the height MAE by 0.22 - 4.33 and the height RMSE by 0.51 - 7.60 in comparison to SARPN \cite{chen2019structure}, DORN \cite{fu2018deep}, and LOFT-FOA+H.
It's worth noting that SARPN \cite{chen2019structure}, DORN \cite{fu2018deep} predicts pixel-wise building height, and MSL-BRN predicts instance-wise building height.
As far as we know, MSL-BRN is the first-ever method to predict instance-wise real-world building height.
Thus, we add a building height head directly to LOFT-FOA \cite{wang2022learning} (\ie LOFT-FOA+H) and compare its prediction results with our own method.
\cref{fig:building-height} presents the qualitative building height prediction results from our method and LOFT-FOA+H.
Regarding the angle prediction tasks, when trained on $BN_{100}$, the PBC module results in an MAE of 9.92 for offset angle prediction and an MAE of 1.22 for off-nadir angle prediction. 
The performance increase demonstrates the efficacy of the PBC, ROFE, and the building height prediction module in a single-level supervised learning scenario.
% 高度指标
\begin{table}[H]
    \centering
    \scalebox{0.8}{
    \begin{tabular}{lcc}
        \toprule
        method                              & height MAE        & height RMSE   \\
        \midrule
        SARPN \cite{chen2019structure}      & 15.23             & 28.69         \\
        DORN \cite{fu2018deep}              & 13.40             & 27.03         \\
        LOFT-FOA+H                          & 11.12             & 21.60         \\
        Ours                                & \textbf{10.90}    & \textbf{21.09}\\
        \bottomrule
    \end{tabular}
    }
    \caption{
    %The quantitative comparison of the baselines and our method trained on $OC_{100}$ and tested on the OmniCity-view3 test set in terms of the building height MAE and RMSE. 
    Building height prediction results of different methods in terms of MAE and MSE trained on $OC_{100}$ and tested on the OmniCity-view3 test set.
    }
    \label{tab:height-perf}
\end{table}

% 模块消融实验
\begin{table}[H]
    \centering
    \scalebox{0.8}{
    \begin{tabular}{lcccc}
        \toprule
        method      & F1-score      & Precision     & Recall        & EPE           
        \\
        \midrule
        baseline    & 61.35         & 61.84         & 61.65              & 5.70          
        \\
        +PBC        & 62.32          &  62.28     &  62.35             & 5.53          
        \\
        +ROFE        & 62.87          & 63.89	       & 62.15          & 5.63          
        \\
        +PBC+ROFE    & \textbf{65.40} & \textbf{66.74}   & \textbf{64.12}              & \textbf{5.49} 
        \\         
        \bottomrule
    \end{tabular}
    }
    \caption{
    %The quantitative comparison of various modules in terms of the footprint segmentation and the offset prediction performance.
    Footprint segmentation results of different modules in terms of F1-score, precision, recall (\%) and offset prediction results in terms of EPE.
    }
    \label{tab:module-ablation}
\end{table}

\begin{figure*}
    \centering
    \includegraphics[width=0.83\linewidth]{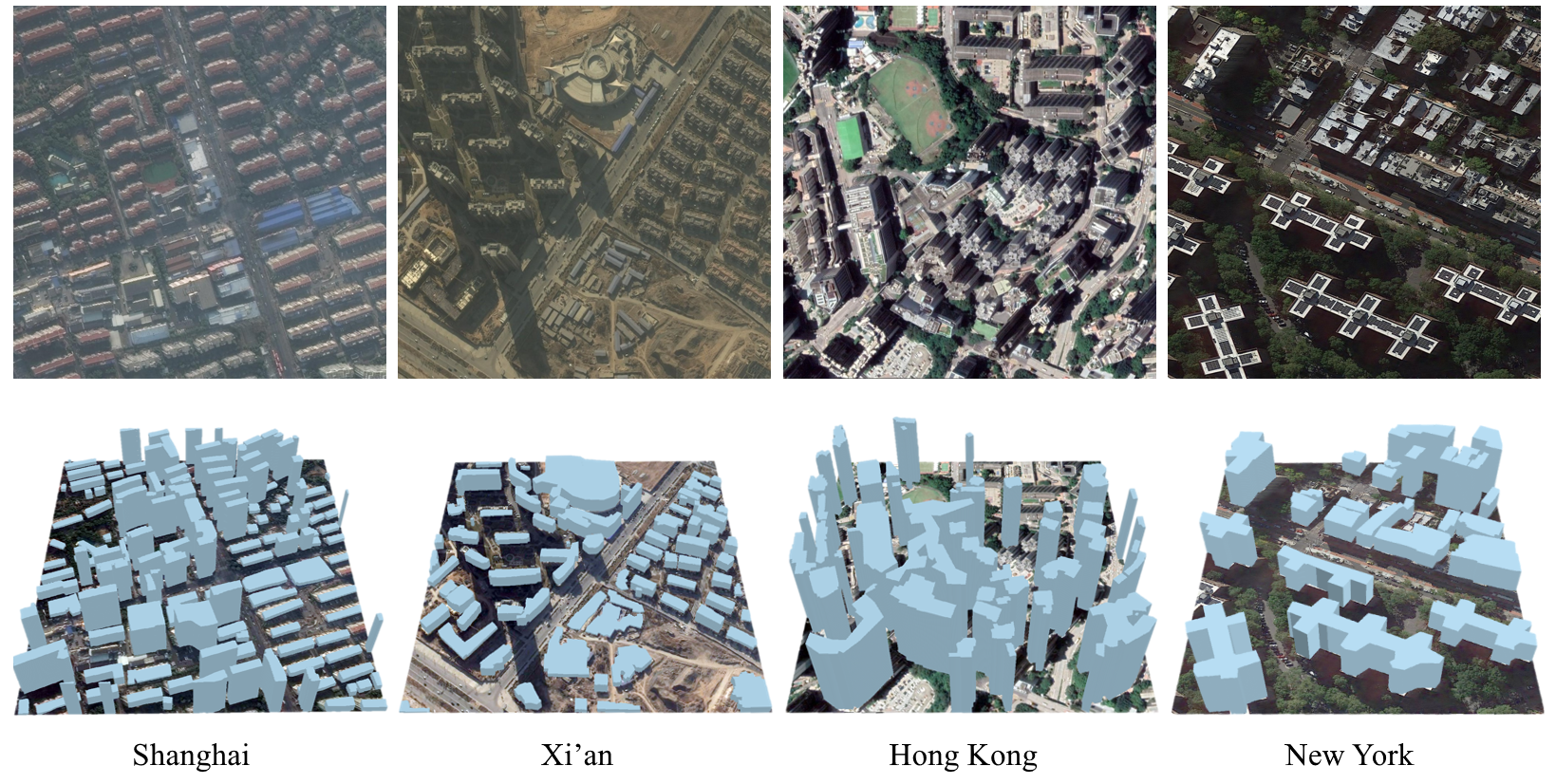}
    \caption{
    3D reconstruction results of Shanghai, Xi'an, Hong Kong, and New York obtained using our method.
    The remote sensing images for Shanghai and Xi'an are chosen from the BONAI test set, whereas the remote sensing image for New York is chosen from the OmniCity-view3 test set.
    }
    \label{fig:3d-recon}
\end{figure*}

% ======================================== Ablation Study =========================================

\subsection{Ablation study}
In this section, we examine the impact of the principal new components of our method: (1) the PBC module; (2) the ROFE module; and (3) the building height head.
Additionally, we will analyze the outcome of the data ablation experiment in the multi-level supervised learning setting.

\textbf{Module ablation.} The outcomes acquired by implementing the aforementioned modules successively on $BN_{30/70}$ are detailed in \cref{tab:module-ablation}. The table provides information on F1-score for footprint segmentation and EPE for offset prediction. 
LOFT-FOA \cite{wang2022learning} is trained on $BN_{30}$ and serves as the baseline.
The second row (+PBC) illustrates the results obtained by applying the PBC module to LOFT-FOA \cite{wang2022learning}.
The results indicate that incorporating the two-angle prediction tasks enhances the F1-score of the footprint extraction by 0.97\%. 
It should be noted that the added offset-unknown 70\% samples in $BN_{30/70}$, which lacks angle ground truth, does not contribute to PBC's training. 
The third row (+ROFE) displays the outcomes achieved by applying the ROFE module to LOFT-FOA \cite{wang2022learning}.
Results demonstrate that, compared with the baseline, prediction of the footprint segmentation guided by predicted offset and roof, coupled with additional 70\% offset-unknown samples from $BN_{30/70}$, leads to a 1.52\% improvement in the F1-score. 
The fourth row (+PBC+ROFE) indicates that the simultaneous inclusion of the PBC and ROFE modules can improve the F1-score of the footprint extraction by 4.05\%. 
The aforementioned results show that PBC and ROFE modules can help to enhance the accuracy of footprint segmentation and offset prediction.

\textbf{Data ablation.} The outcomes of our approach trained on various dataset combinations concerning F1-score for footprint segmentation, and EPE for offset prediction are shown in \cref{tab:data-ablation}. 
The first line ($\mathcal{X}^{OH}$) displays the results of training LOFT-FOA \cite{wang2022learning} on 30\% of OmniCity-view3 \cite{li2023omnicity} $\mathcal{X}^{OH}$ samples ($OC_{30}$).
The second row ($\mathcal{X}^{OH}$+$\mathcal{X}^{H}$) shows the results of our method trained on a mix of 30\% of OmniCity-view3 \cite{li2023omnicity} $\mathcal{X}^{OH}$ samples ($OC_{30}$) and 30\% of the OmniCity-view3 $\mathcal{X}^{H}$ samples.
The results demonstrate a 3.28\% improvement in F1-score for footprint extraction compared to LOFT-FOA \cite{wang2022learning} trained solely on $OC_{30}$.
The third row ($\mathcal{X}^{OH}$+$\mathcal{X}^{H}$+$\mathcal{X}^{N}$) presents the outcomes of our methodology, trained on a mix of 30\% of OmniCity-view3 \cite{li2023omnicity} $\mathcal{X}^{OH}$ samples, 30\% of OmniCity-view3 \cite{li2023omnicity} $\mathcal{X}^{H}$ samples, and the rest 40\% of OmniCity-view3 \cite{li2023omnicity} $\mathcal{X}^{N}$ samples.
The results demonstrate a 0.44\% increase in F1-score compared to our method trained on $\mathcal{X}^{OH}$+$\mathcal{X}^{H}$, indicating the effectiveness of including $\mathcal{X}^{N}$ samples.
The reason for training LOFT-FOA \cite{wang2022learning} instead of our method on $OC_{30}$ (first row) is to evaluate the gain in a scenario where $\mathcal{X}^{H}$ and $\mathcal{X}^{N}$ samples are available by using our method.

% 数据消融实验
\begin{table}[H]
    \centering
    \scalebox{0.8}{
    \begin{tabular}{lcccc}
        \toprule
        data                                                    & F1            & Precision     & Recall        & EPE
        \\
        \midrule
        $\mathcal{X}^{OH}$                                      & 67.09         & 63.23         & 71.47         & 6.08              
        \\
        %$\mathcal{X}^{OH}$                                      & 70.45         & ?????         & ?????         & 6.08              \\
        $\mathcal{X}^{OH}$+$\mathcal{X}^{H}$                    & 70.37         & 65.35         & \textbf{76.24}         & 5.99              
        \\
        $\mathcal{X}^{OH}$+$\mathcal{X}^{H}$+$\mathcal{X}^{N}$  & \textbf{70.81}& \textbf{66.15}& 76.18         & \textbf{5.84}              
        \\
        \bottomrule
    \end{tabular}
    }
    \caption{
    %The quantitative comparison of the baseline and our method trained on different dataset combinations in terms of the footprint segmentation and the offset prediction performance. 
    Building footprint segmentation results of different methods in terms of F1-score, precision, recall (\%) and offset prediction results in terms of EPE trained on different dataset combinations.
    }
    \label{tab:data-ablation}
\end{table}

% ====================================== 3D Reconstruction ========================================
\subsection{3D reconstruction results of different cities}
\cref{fig:3d-recon} shows the 3D reconstruction results of four cities (\ie Shanghai, Xi'an, Hong Kong, and New York) obtained from our method.
The results demonstrate the effectiveness of our method on 3D building reconstruction across different cities.
Note that we use the method in \cite{zorzi2021machine} to regularize the predicted building footprint masks.

%% file: sec/5_conclusion.tex
\section{Conclusion}
\label{sec:conclusion}

%-------------------------------------------------------------------------

In this paper, we have presented a new method for multi-level supervised building reconstruction from monocular remote sensing images, which is capable of reconstructing the accurate 3D building models using samples of different annotation levels.
Qualitative and quantitative evaluations confirm that our method achieves competitive performance and significantly enhances the 3D building reconstruction capability in comparison to the current state-of-the-art across diverse experimental settings.
The effect of the Pseudo Building Bbox Calculator and the Roof-Offset guided Footprint Extractor, as well as the annotation levels of the samples were also analyzed in the ablation study.
Furthermore, we expanded the monocular building reconstruction datasets to encompass additional cities.
We believe that our approach offers efficient and cost-effective solutions for 3D building reconstruction in complex real-world scenes. %and significantly improves the exploitation of current remote sensing data.
In our future work, we would like to investigate more effective strategies to improve the 3D building reconstruction performance whilst exploring more adaptable and practical techniques for large-scale city modeling.

% \begin{flushleft}
% \textbf{Acknowledgements.} 
% % This project was funded in part by National Natural Science Foundation of China (No. 42201358 and No. 62325111), National Key R\&D Program of China (No. 2022ZD0160101) and Shanghai Artificial Intelligence Laboratory.
% This project was funded in part by National Natural Science Foundation of China (Grant No. 42201358 and No. 62325111) and Shanghai Artificial Intelligence Laboratory.
% \end{flushleft}

\par

\noindent \textbf{Acknowledgements.} 
This project was funded in part by National Natural Science Foundation of China (Grant No. 42201358 and No. 62325111) and Shanghai Artificial Intelligence Laboratory.

%% file: sec/6_suppl.tex
\clearpage
\setcounter{page}{1}
\maketitlesupplementary

In this supplementary material, we first provide additional details of our proposed MLS-BRN
model (\cref{sec:sup-methods}).
Then we provide additional details of our newly released datasets, as well as the sample diversity used in our study (\cref{sec:sup-data}).
Last, we provide additional experimental results of building footprint segmentation, offset angle prediction, and 3D building reconstruction (\cref{sec:sup-exp}).

% \begin{figure}[H]
%     \centering
%     \includegraphics[width=1.0\linewidth]{figures/sup-framework.png}
%     \caption{
%     The utilization details of the ground truth of samples with different supervision levels.
%     The green dotted lines denote the supervision of the off-nadir angle head and offset angle head in PBC.
%     The blue dotted lines denote the calculation of the pseudo building bbox of $\mathcal{X}^{OH}$.
%     The orange dotted line denotes the calculation of the pseudo building bbox of $\mathcal{X}^{N}$.
%     }
%     \label{fig:sup-framework}
% \end{figure}

% ======================================= Methods ================================================、
\appendix
\section{Additional details of methods}
\label{sec:sup-methods}

\subsection{Additional training details}
\begin{figure}[H]
    \centering
    \includegraphics[width=1.0\linewidth]{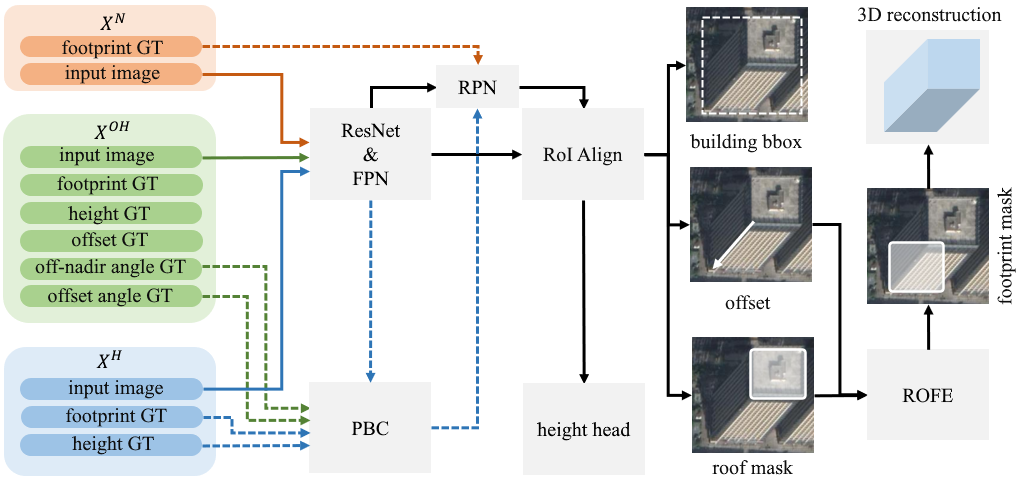}
    \caption{
    The utilization details of the ground truth of samples with different supervision levels.
    The green dotted lines indicate the supervision of the off-nadir angle head and offset angle head in PBC using the ground truth provided by $\mathcal{X}^{OH}$.
    The blue dotted lines denote the calculation of the pseudo building bbox of $\mathcal{X}^{H}$.
    The orange dotted line denotes the calculation of the pseudo building bbox of $\mathcal{X}^{N}$.
    }
    \label{fig:sup-framework}
\end{figure}

In our proposed model, different levels of samples are supervised with different
training strategies. 
Consequently, the ground truth of different levels of samples is utilized differently (\cref{fig:sup-framework}).
The PBC module employs the building footprint and height ground truth of $\mathcal{X}^{H}$ to compute the pseudo building bboxes, while the building footprint and height ground truth of $\mathcal{X}^{OH}$ are not used by PBC since their building bbox ground truth is already known.
However, PBC uses the off-nadir angle and offset angle ground truth of $\mathcal{X}^{OH}$ for supervising the training of the two angle heads.
Furthermore, PBC cannot calculate the pseudo building bboxes for $\mathcal{X}^{N}$ since they have no building height ground truth. 
Instead, the pseudo building bboxes of $\mathcal{X}^N$ are calculated by enlarging the building footprint ground truth by a certain percentage.
%%lwj: 这个补充的根本不行. 还big enough...

\subsection{Additional implementation details}
In our proposed model, the feature map sent to PBC for calculating the pseudo building bbox is the largest layer from FPN (\ie the layer with the size of 256 $\times$ 256). 
The off-nadir angle head of PBC is composed of 4 Conv layers and 3 FC layers, while the off-nadir angle head of PBC is composed of 8 Conv layers and 6 FC layers.

\subsection{Additional details of 3D model reconstruction}
We apply the method outlined in \cite{zorzi2021machine} to regularize the predicted building footprint mask obtained from our MLS-BRN. 
Subsequently, we use the Douglas–Peucker algorithm \cite{douglas1973algorithms} to simplify the regularized polygons by reducing extraneous vertices. 
Furthermore, the raster polygons are converted to vector data format for visualization.
Lastly, the vectorized polygons are combined with the predicted building height to complete the 3D building reconstruction.

% ======================================= Datasets ================================================
\section{Additional details of datasets}
\label{sec:sup-data}

\subsection{Details of existing building datasets}

\cref{tab:sup-dataset} lists some popular building footprint extraction and 3D reconstruction datasets (with offset or height annotations).
The public building footprint extraction datasets far exceed the 3D reconstruction datasets in terms of both the number of images and the number of building instances. %%lwj-done:samples和annotations分别指的是表格里的什么？不明确。
Our MLS-BRN demonstrates the great potential of leveraging large-scale footprint segmentation datasets to improve 3D building reconstruction performance and reduce the need for 3D annotations.

%%lwj-done: Annotaion按我们的三种监督分三列，数量都用多少K来表示，后面几列用打勾打叉，第一列数据集要有引用。参考下面omnicity论文表格。
\begin{table}[H]
\resizebox{.47\textwidth}{!}{
%\scriptsize
    \centering
    \begin{tabular}{lcccccc}
        \toprule
        Dataset             & \#Images      & \#Instances   & Off-Nadir     & Foot.     & Offset        & Height
        \\
        \midrule
        Microsoft \cite{microsoft}    & -             & 1,240M        & $\times$      & \checkmark    & $\times$      & $\times$
        \\
        Open Bld. \cite{sirko2021continental}      & -             & 1,800M        & $\times$      & \checkmark    & $\times$      & $\times$ 
        \\
        CrowdAI \cite{2018Crowdai}     & 340K          & 2,915K        & $\times$      & \checkmark    & $\times$      & $\times$ 
        \\
        WHU \cite{ji2019building}     & 8.2K          & 120K          & $\times$      & \checkmark    & $\times$      & $\times$
        \\
        SpaceNet  \cite{2018DeepGlobe}   & 24.6K         & 303K          & $\times$      & \checkmark    & $\times$      & $\times$ 
        \\
        MVOI \cite{weir2019spacenet}  & 60K           & 127K          & \checkmark    & \checkmark    & $\times$      & $\times$
        \\
        OmniCity \cite{li2023omnicity}           & 75K           & 2,573K        & \checkmark    & \checkmark    & $\times$      & \checkmark
        \\
        DFC19  \cite{2020Learning}      & 3.2K          & 500K          & \checkmark    & \checkmark    & \checkmark    & \checkmark
        \\
        ATL-SN4 \cite{2020Learning}    & 8K            & 1,100K        & \checkmark    & \checkmark    & \checkmark    & \checkmark
        \\
        BONAI  \cite{wang2022learning}  & 3.3K          & 269K          & \checkmark    & \checkmark    & \checkmark    & \checkmark
        \\
        ISPRS 3D \cite{ISPRS}  & 0.033K        & -             & \checkmark    & \checkmark    & \checkmark    & \checkmark
        \\
        \bottomrule
    \end{tabular}
    }
    \caption{
    A summary of popular building footprint segmentation and 3D reconstruction datasets.
    Foot. is the abbreviation for footprint.
    }
    \label{tab:sup-dataset}
\end{table}

\begin{figure*}
    \centering
    \includegraphics[width=0.95\linewidth]{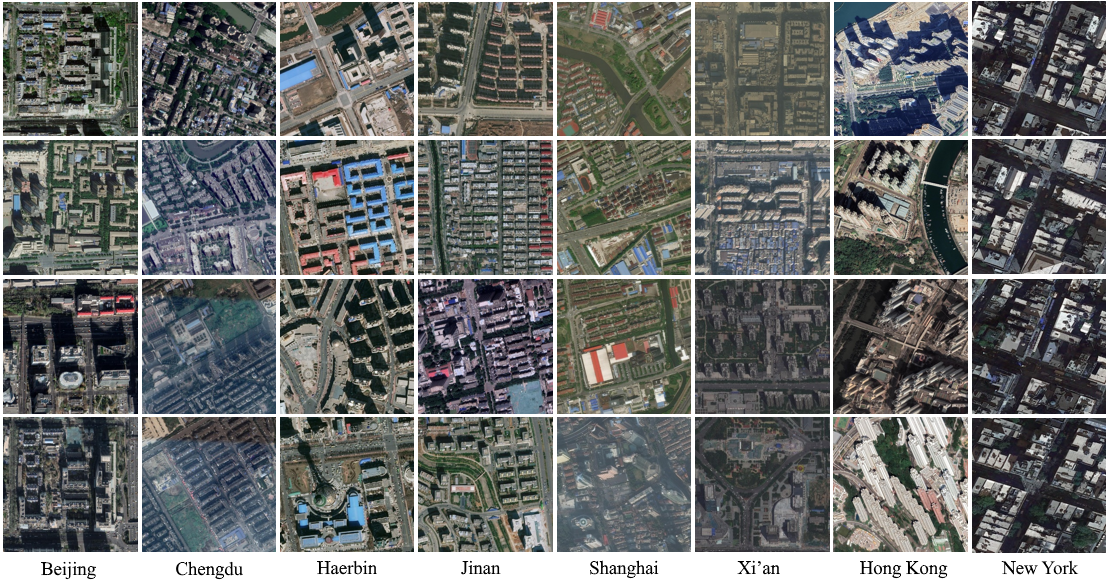}
    \caption{
    Remote sensing images of 8 cities. 
    The remote sensing images of Beijing, Chengdu, Harbin, Jinan, Shanghai and Xi’an are chosen from the BONAI dataset. The images of New York are chosen from the OmniCity-view3 dataset. The images of Hong Kong are chosen from the HK dataset.
    }
    \label{fig:sup-sample}
\end{figure*}

\subsection{Details of samples of each city}

In \cref{fig:sup-sample}, we provide some examples of the remote sensing image samples used in our datasets, which demonstrate a high diversity of each city in terms of the off-nadir angle, offset angle, as well as the building density, areas, height, etc.

\subsection{Additional details of newly released dataset}
\begin{figure}[H]
    \centering
    \includegraphics[width=0.85\linewidth]{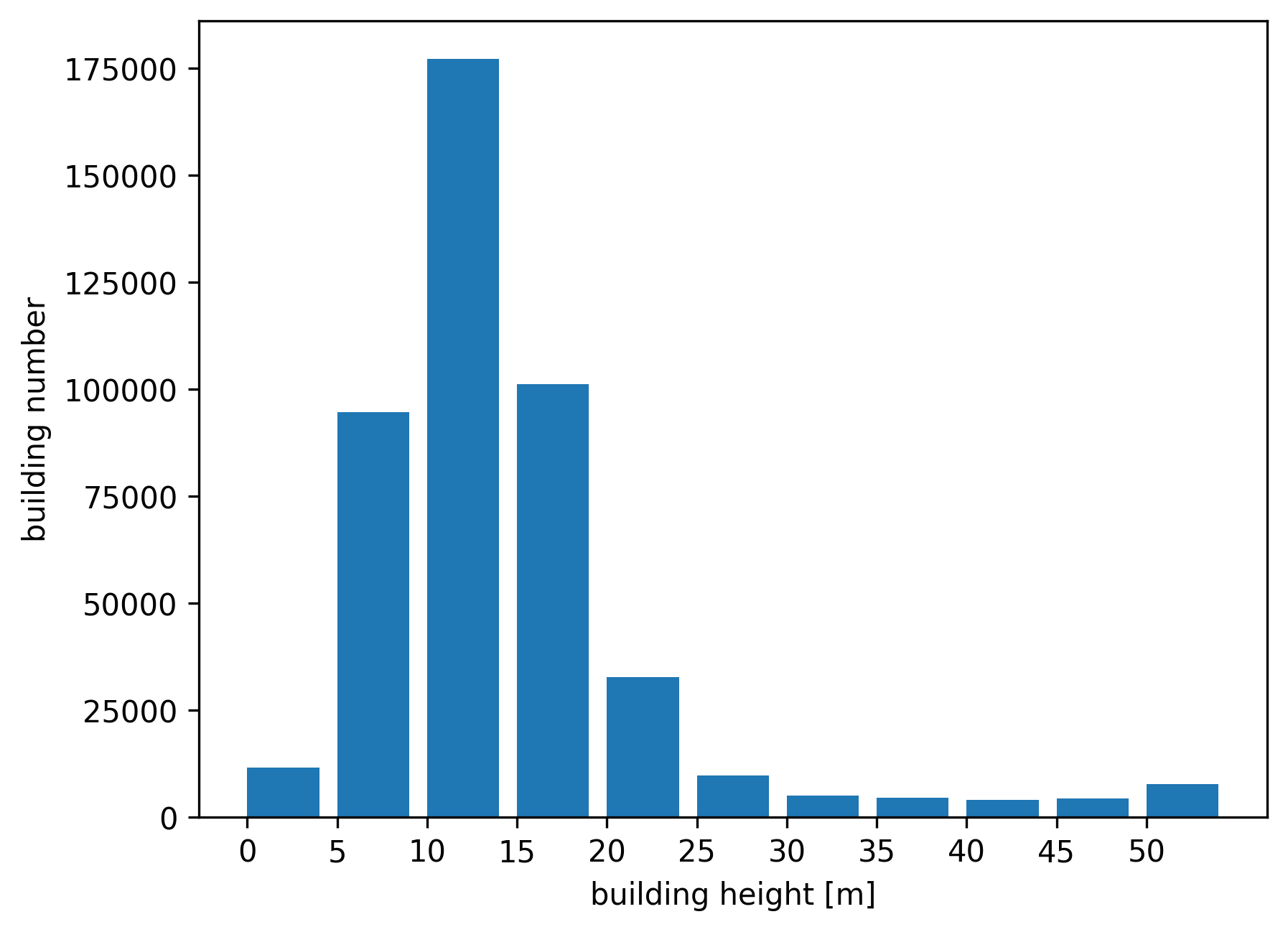}
    \caption{
    The building height distribution of OmniCity-view3.
    }
    \label{fig:sup-height-count-oc}
\end{figure}
\begin{figure}[H]
    \centering
    \includegraphics[width=0.85\linewidth]{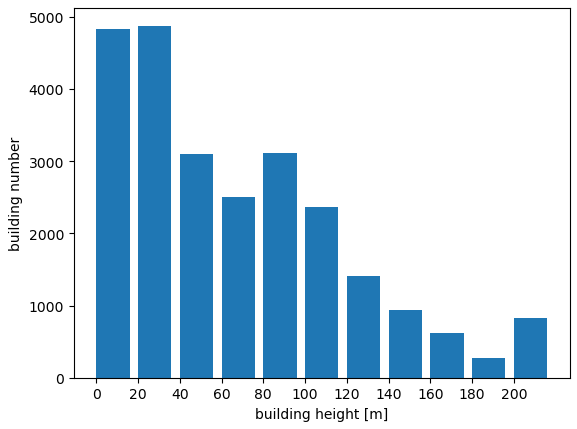}
    \caption{
    The building height distribution of HK.
    }
    \label{fig:sup-height-count-hk}
\end{figure}

In this study, we provide additional offset annotations for the view3 subset of OmniCity (denoted by OmniCity-view3) since this subset contains images with the largest off-nadir angles.
Specifically, we annotate roof-to-footprint offsets for 17,092 and 4,929 images from trainval and test sets, respectively. \cref{fig:sup-height-count-oc} demonstrates the building height distribution of OmniCity-view3 dataset.
We also release a new dataset collected from Hong Kong (denoted by HK dataset), containing 500 remote sensing images for the trainval set and 119 images for the test set, all of which are annotated with building footprint, roof-to-footprint offset, and building height.
The remote sensing images are cropped to 1024 $\times$ 1024 and contain 24,851 annotated buildings in total.
\cref{fig:sup-height-count-hk} demonstrates the building height distribution of HK dataset.

% ==================================== Experiments ===============================================
\section{Additional experimental results}
\label{sec:sup-exp}

\subsection{Ablation study on multi-level sample division}

% \cref{tab:sup-sample-division} displays the footprint segmentation and offset prediction performance of our method trained on datasets with different proportions of $\mathcal{X}^{OH}$ and $\mathcal{X}^{H}$ samples. The performance of their corresponding experiments only trained on the $\mathcal{X}^{OH}$ are also listed.
% The results show that with the increase in the proportion of $\mathcal{X}^{OH}$, the footprint segmentation and offset prediction performance is getting better.
\cref{tab:sup-sample-division} displays the footprint segmentation and offset prediction performance of our method trained on datasets with different proportions of $\mathcal{X}^{OH}$ and $\mathcal{X}^{H}$ samples. 
The performance of LOFT-FOA \cite{wang2022learning} trained only on the $\mathcal{X}^{OH}$ samples are also listed for better demonstrating the 
performance gains from introducing different percentages of $\mathcal{X}^{H}$ samples.
The results show that the building footprint segmentation performance difference between LOFT-FOA \cite{wang2022learning} and our method is getting smaller with the increase in the proportion of $\mathcal{X}^{OH}$ samples.
In the main paper, we opt for the ratio of 30\%:70\% since the building footprint performance of our method, trained on $BN_{30/70}$, surpasses that of LOFT-FOA \cite{wang2022learning} trained on $BN_{100}$.

% \begin{figure}[H]
%     \centering
%     \includegraphics[width=0.9\linewidth]{figures/sup-footprint-method-nyhk.png}
%     \caption{
%     The footprint segmentation results of New York (the first two rows) and Hong Kong (the last two rows) from baselines and our method trained on $OC_{100}$ and $BH_{100}$, respectively.
%     The images demonstrate high diversity in terms of building layout.
%     }
%     \label{fig:sup-footprint-method-nyhk}
% \end{figure}
\begin{table}[H]
\footnotesize
    \centering
    \begin{tabular}{lccccc}
        \toprule
        dataset     & Model     & F1        & Precision & Recall    & EPE           
        \\
        \midrule
        $BN_{10}$   &LOFT-FOA   & 53.91     & 53.28     & 54.55     & 7.42
        \\
        $BN_{10/90}$&Ours       & 63.18     & 65.05     & 61.42     & 6.14
        \\
        \midrule
        $BN_{20}$   &LOFT-FOA   & 59.65     & 59.05     & 60.27     & 5.79  
        \\
        $BN_{20/80}$&Ours       & 64.47     & 67.71     & 61.52     & 5.49  
        \\
        \midrule
        $BN_{30}$   &LOFT-FOA   & 61.35     & 61.84	    & 61.65	    & 5.70
        \\
        $BN_{30/70}$&Ours       & 65.50     & 66.94     & 64.11     & 5.39 
        \\
        \midrule
        $BN_{40}$   &LOFT-FOA   & 63.17     & 62.79     & 63.56     & 5.26
        \\
        $BN_{40/60}$&Ours       & 65.78     & 66.16     & 65.40     & 5.22 
        \\
        \midrule
        $BN_{100}$  &LOFT-FOA   & 64.31     & 63.37     & 65.29     & 4.94
        \\
        $BN_{100}$  &Ours       & 66.36     & 65.90     & 66.83     & 4.76
        \\
        \bottomrule
    \end{tabular}
    \caption{
    The experimental results of datasets with different proportions of $\mathcal{X}^{OH}$ and $\mathcal{X}^{H}$ samples.
    As described in the main paper, $BN_{x/y}$ means x\% of BONAI trainval samples are of $\mathcal{X}^{OH}$ type and y\% are of $\mathcal{X}^{H}$ type.
    The results of LOFT-FOA and our method trained on $BN_{100}$ are also listed for better comparison with our methods trained on datasets composed of both $\mathcal{X}^{OH}$ and $\mathcal{X}^{H}$ samples.
    }
    \label{tab:sup-sample-division}
\end{table}
\subsection{Additional results on footprint segmentation}
%\subsubsection{Single-level supervised learning results}
%\subsubsection{Multi-level supervised learning results}

\begin{figure}[H]
    \centering
    \includegraphics[width=0.88\linewidth]{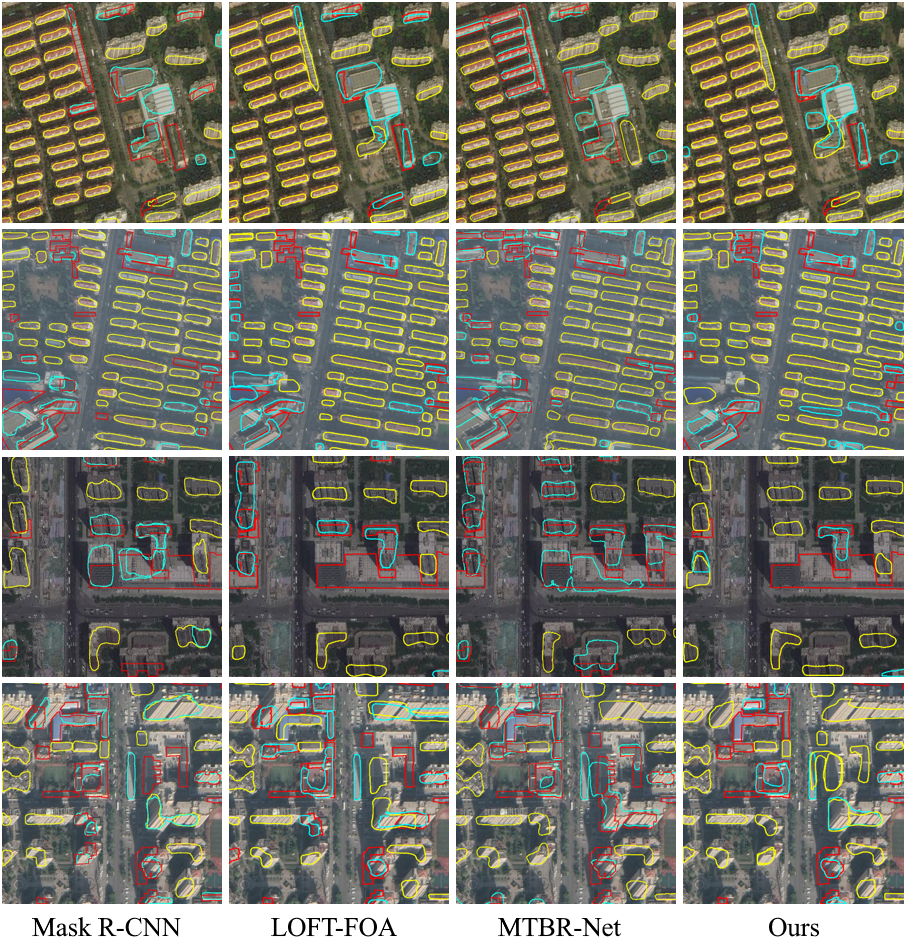}
    \caption{
    The footprint segmentation results of Shanghai and Xi'an from models trained on $BN_{100}$.
    The first two rows display the results of Shanghai, and the last two rows display the results of Xi'an.
    }
    \label{fig:sup-footprint-method-shxa}
\end{figure}

\begin{figure}[H]
    \centering
    \includegraphics[width=0.88\linewidth]{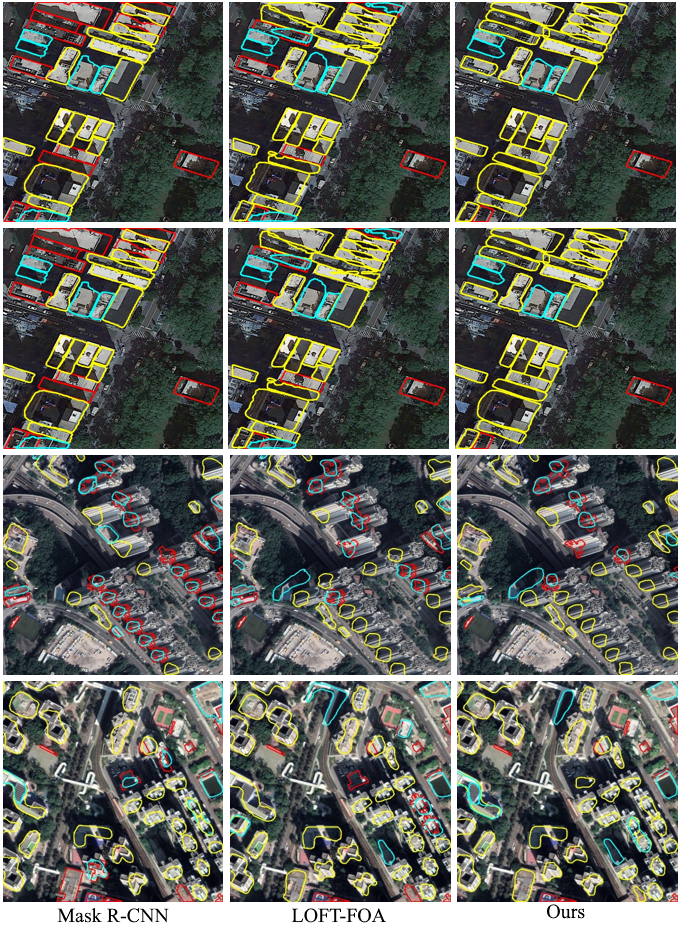}
    \caption{
    The footprint segmentation results of different models trained on $OC_{100}$ and $BH_{100}$, respectively.
    The first two rows display the results of New York (OmniCity-view3), and the last two rows display the results of Hong Kong (HK dataset).
    }
    \label{fig:sup-footprint-method-nyhk}
\end{figure}

\begin{figure}[H]
    \centering
    \includegraphics[width=0.95\linewidth]{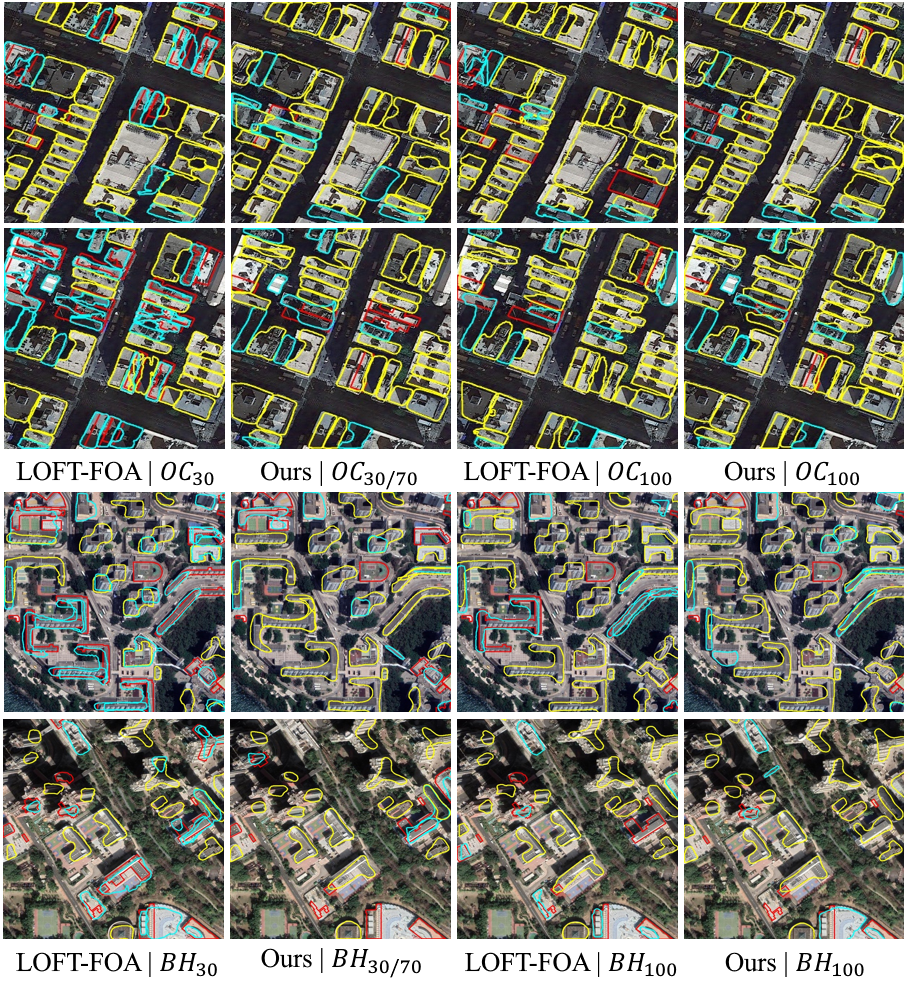}
    \caption{
    The footprint segmentation results of New York (the first two rows) and Hong Kong (the last two rows) from LOFT-FOA and our method trained on $OC_{x}$ and $BH_{x}$, respectively.
    Note that LOFT-FOA$|OC_{30}$ means the results of LOFT-FOA trained on $OC_{30}$.
    }
    \label{fig:sup-footprint-dataset-nyhk}
\end{figure}

\cref{fig:sup-footprint-method-shxa} and \cref{fig:sup-footprint-method-nyhk} demonstrate the additional building footprint segmentation results of four different cities (\ie Shanghai, Xi'an, New York, and Hong Kong) from different models trained on solely $\mathcal{X}^{OH}$ samples.
\cref{fig:sup-footprint-dataset-nyhk} display the building footprint segmentation results of two different cities (\ie New York and Hong Kong) from LOFT-FOA \cite{wang2022learning} and our method trained on datasets containing $\mathcal{X}^{OH}$ and $\mathcal{X}^{H}$ samples.


% \begin{figure*}
%     \centering
%     \includegraphics[width=1.0\linewidth]{figures/sup-offset-angle-bonai.png}
%     \caption{
%     The offset angle prediction results of Shanghai (the first row) and Xi'an (the second row).
%     The red line with the arrow denotes the offset angle ground truth, while the blue line with the arrow denotes the predicted offset angle.
%     }
%     \label{fig:sup-offset-angle}
% \end{figure*}

\subsection{Additional offset angle prediction results}

\begin{figure}[H]
    \centering
    \includegraphics[width=1.0\linewidth]{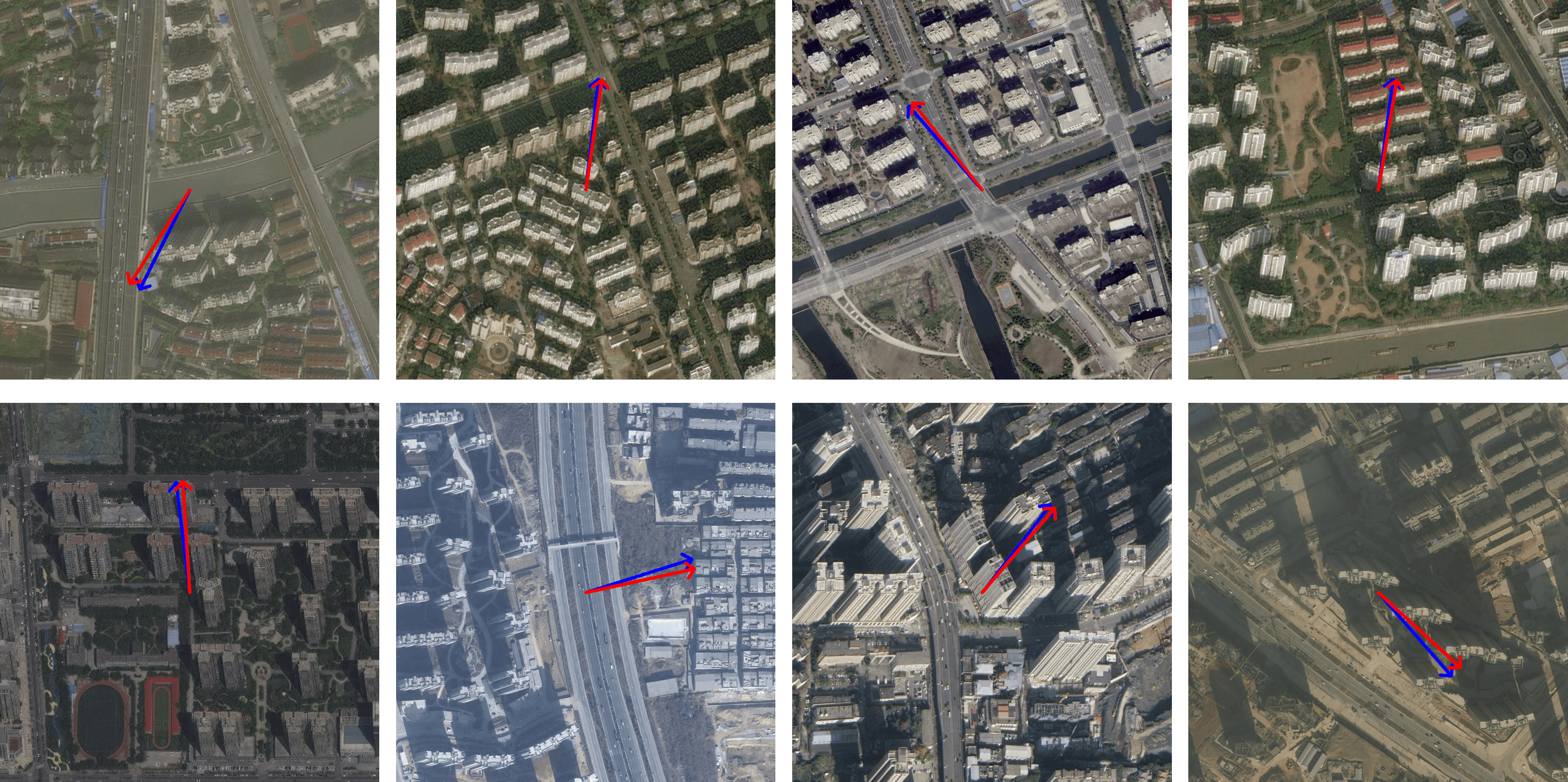}
    \caption{
    The offset angle prediction results of Shanghai (the first row) and Xi'an (the second row).
    The red line with the arrow denotes the offset angle ground truth, while the blue line with the arrow denotes the predicted offset angle.
    }
    \label{fig:sup-offset-angle}
\end{figure}

\cref{fig:sup-offset-angle} demonstrates the offset angle prediction results of our method.
To aid comprehension, a vector is used to represent the offset angle, with the vector direction pointing from the footprint to the roof. For example, a vector pointing horizontally to the right denotes a 0 degree angle, whereas a vector pointing downwards vertically denotes a 90 degree angle.

%\subsection{Bad cases}
\subsection{Failure case analysis}
\begin{figure}[H]
    \centering
    \includegraphics[width=1.0\linewidth]{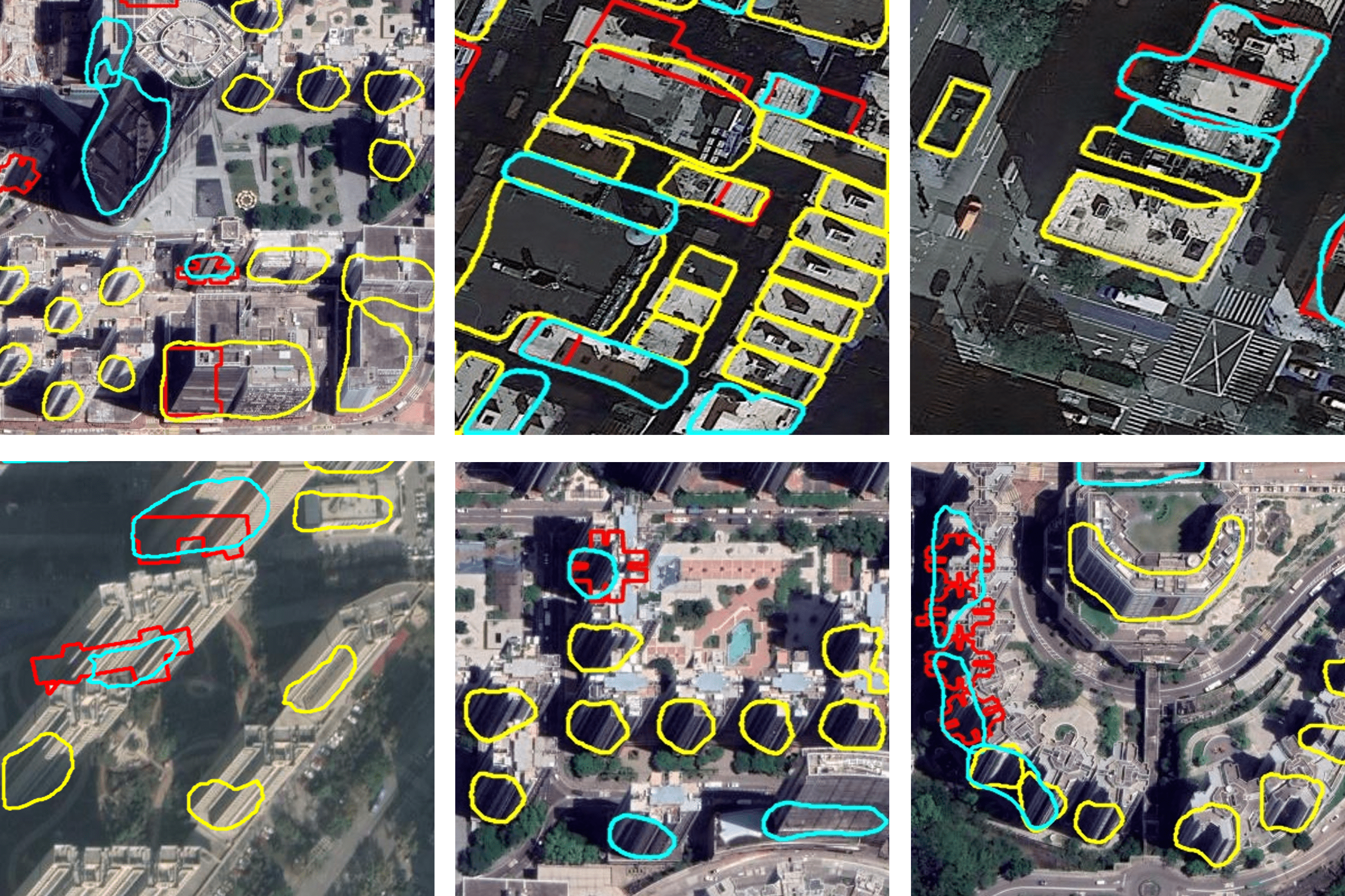}
    \caption{
    Some typical failures of footprint segmentation results.
    The yellow, cyan, and red polygons denote the TP, FP, and FN.
    }
    \label{fig:sup-bad-cases}
\end{figure}

%\cref{fig:sup-bad-cases} displays some typical building footprint segmentation failure cases.
\cref{fig:sup-bad-cases} displays some typical failure cases obtained from our method.
The most common failure cases include: (1) the mixing up of the building roof and facade (the first column); (2) inaccurate segmentation of a complex building roof (the second column); and (3) the misinterpretation of multiple roofs as one roof, or vice versa (the third column).

\subsection{Additional 3D building reconstruction results}

% \cref{fig:sup-recon} shows more 3D reconstruction results of different cities from our method as well as their corresponding ground truth.
% \cref{fig:sup-outdomain-recon} shows the 3D reconstruction results of cities absent from the corresponding train set.
% The results show that our model has a good generalization performance in terms of 3D building reconstruction tasks.

\begin{figure*}
    \centering
    \includegraphics[width=1.0\linewidth]{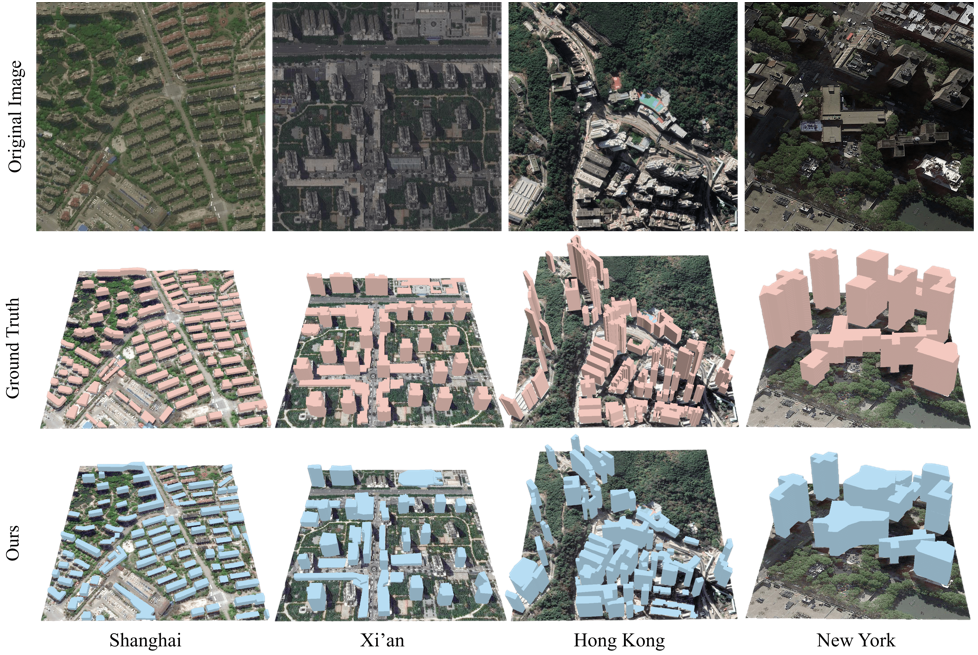}
    \caption{
    The 3D reconstruction results of Shanghai, Xi'an, Hong Kong, and New York.
    }
    \label{fig:sup-recon}
\end{figure*}

\begin{figure*}
    \centering
    \includegraphics[width=1.0\linewidth]{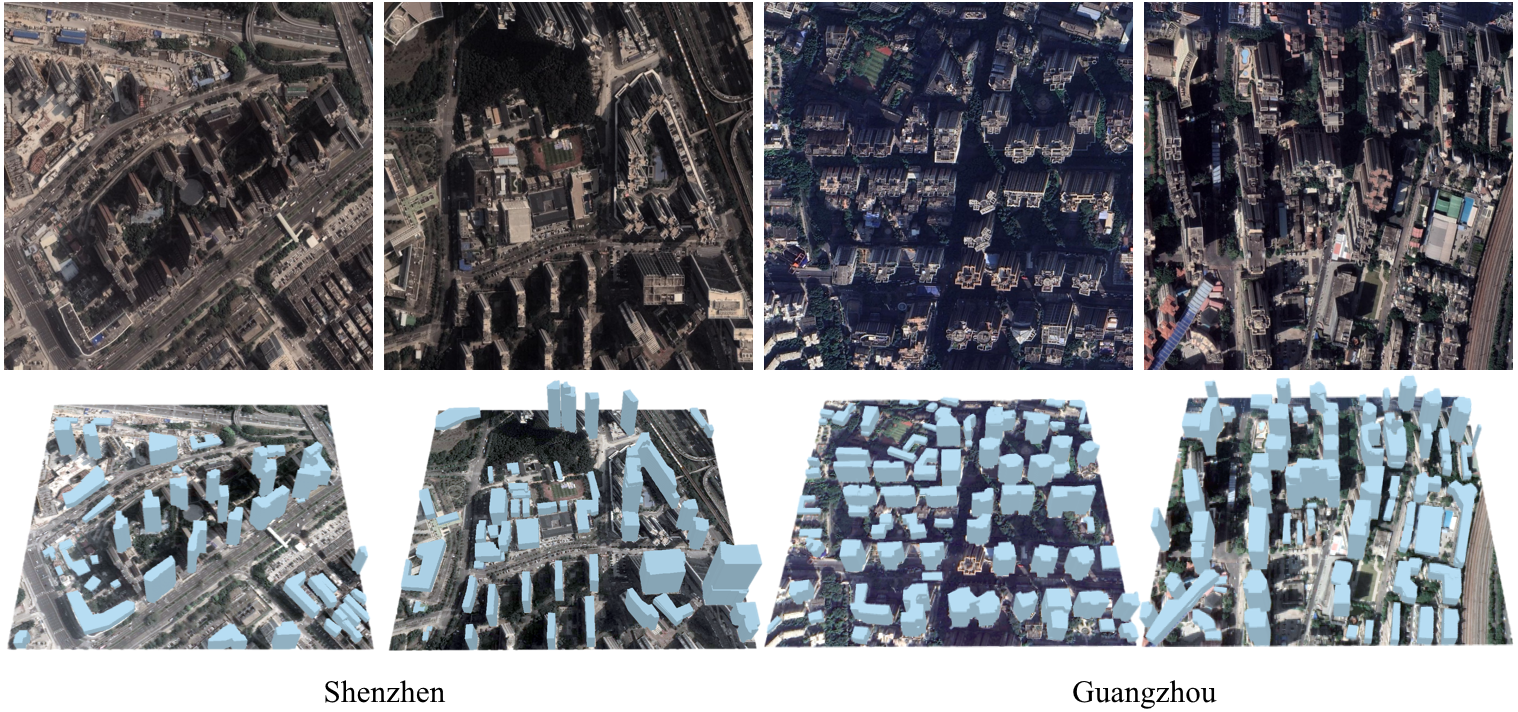}
    \caption{
    The 3D reconstruction results of Shenzhen and Guangzhou.
    }
    \label{fig:sup-outdomain-recon}
\end{figure*}

\cref{fig:sup-recon} shows additional 3D reconstruction results of four different cities from our method, alongside their corresponding ground truth. Moreover, in order to demonstrate the generalization performance of our method in new regions, \cref{fig:sup-outdomain-recon} shows the 3D reconstruction results of two new cities, i.e., Shenzhen and Guangzhou. The results indicate that our model has a good generalization performance in terms of 3D building reconstruction tasks.

\clearpage

